\documentclass[10pt,journal,letterpaper,final,twoside]{IEEEtran}
\usepackage{amsmath, amsthm, amssymb}
\usepackage{amssymb}
\usepackage{url,flushend,multirow}
\usepackage{algorithm,algorithmic}
\usepackage[dvips]{graphicx}
\usepackage{bm}
\usepackage{float,subfigure}
\usepackage[nocompress]{cite}
\newcommand{\tr}{\textrm{Tr}}
\renewcommand{\algorithmicrequire}{\textbf{Input:}}   
\renewcommand{\algorithmicensure}{\textbf{Output:}}  

\begin{document}
%
\title{Relational Multi-Manifold Co-Clustering}

\author{Ping~Li,~
        Jiajun~Bu,~\IEEEmembership{Member,~IEEE,}
        Chun~Chen,~\IEEEmembership{Member,~IEEE,}
        Zhanying~He,
        and~Deng~Cai,~\IEEEmembership{Member,~IEEE}
\thanks{Manuscript received August 7, 2012; revised November 25, 2012; accepted December 10, 2012.}
\thanks{P. Li, J. Bu, C. Chen and Z. He are with the Zhejiang Provincial Key Laboratory of Service Robot, College of Computer Science, Zhejiang University, Hangzhou 310027, China (e-mail: \{lpcs, bjj, chenc, hezhanying\}@zju.edu.cn).}
\thanks{D. Cai is with the State Key Lab of CAD\&CG, College of Computer Science, Zhejiang University,
Hangzhou 310058, China (e-mail: dengcai@cad.zju.edu.cn)}
}

%
%

\markboth{For personal use only}
{LI \MakeLowercase{\textit{et al.}}:~RELATIONAL MULTI-MANIFOLD CO-CLUSTERING}

\maketitle

\begin{abstract}
Co-clustering targets on grouping the samples (\textit{e.g.}, documents, users) and the features (\textit{e.g.}, words, ratings) simultaneously. It employs the dual relation and the bilateral information between the samples and features. In many real-world applications, data usually reside on a submanifold of the ambient Euclidean space, but it is nontrivial to estimate the intrinsic manifold of the data space in a principled way. In this study, we focus on improving the co-clustering performance via manifold ensemble learning, which is able to maximally approximate the intrinsic manifolds of both the sample and feature spaces. To achieve this, we develop a novel co-clustering algorithm called \emph{Relational Multi-manifold Co-clustering} (RMC) based on symmetric nonnegative matrix tri-factorization, which decomposes the relational data matrix into three submatrices. This method considers the inter-type relationship revealed by the relational data matrix, and also the intra-type information reflected by the affinity matrices encoded on the sample and feature data distributions. Specifically, we assume the intrinsic manifold of the sample or feature space lies in a convex hull of some pre-defined candidate manifolds. We want to learn a convex combination of them to maximally approach the desired intrinsic manifold. To optimize the objective function, the multiplicative rules are utilized to update the submatrices alternatively. Besides, both the \textit{entropic mirror descent algorithm} and the \textit{coordinate descent algorithm} are exploited to learn the manifold coefficient vector. Extensive experiments on documents, images and gene expression data sets have demonstrated the superiority of the proposed algorithm compared to other well-established methods.
\end{abstract}


%
\IEEEpeerreviewmaketitle

\section{Introduction}
\IEEEPARstart{T}{he} last dozen years have witnessed the profound changes in human lifestyle with the rapid development of modern digital technologies. A huge amount of multi-type relational data emerge every moment in a broad range of real-world applications \cite{chen2007simulcluster,long2006sc-multirelation}, \textit{e.g.}, numerous documents in online offices, various images or videos in the social networks, and gene expression data for medical diagnosis. Clustering has established itself as a very useful tool to handle a vast number of data with successful applications in knowledge management, information retrieval and bioinformatics, \textit{etc}. As an unsupervised learning mechanism, clustering seeks the appropriate partitioning of the data with the rule that the data points within the same cluster should be more closely and mutually interdependent than those in different clusters. In general, traditional clustering belongs to the unilateral learning, namely it only emphasizes clustering along the sample or feature dimension. Recent works have shown that clustering samples and features simultaneously, \textit{i.e.}, co-clustering, is beneficial to further improving the clustering performance, in the sense that co-clustering fully makes use of the dual interdependence between samples and features to discover certain hidden clustering structures in data \cite{dhillon2001coclu-bipartite, dhillon2003coclu-info, sindhwani2008regu-coclu}.

Actually, a great many of important applications require to co-cluster both the samples and features, which are often stacked in columns and rows of a dyadic data matrix, such as co-occurrence matrix, rating matrix and proximity matrix \cite{long2005coclu-bvd}. For instance, in the text and webpage analysis, the relations between words and documents can be reflected by a contingency table \cite{dai2007coclu-doc, pan2008crd}. In a recommendation system, users and ratings on items (\textit{e.g.}, movies, music) can be co-clustered via collaborative filtering \cite{chen2009cf-onmtf}. In the biological domain (\textit{e.g.}, cancer diagnosis, mircoarray analysis), we can cluster the genes and experimental conditions simultaneously \cite{hanisch2002coclu-bio}. To this end, many co-clustering algorithms have emerged, such as graph partitioning based model \cite{dhillon2001coclu-bipartite, rege2006coclu-bipartie,zha2001bipartite}, pattern based model \cite{xu2006pattern-coclu}, information theory based model \cite{dhillon2003coclu-info, slonim2000mutualinfo}, and matrix factorization based model \cite{chen2010nmf-semi-coclu, ding2006onmtf, wang2011fast-nmtf}.

In this work, we focus on matrix factorization based co-clustering, which models the sample-feature relationship from the data reconstruction perspective. A heuristic method is to use singular value decomposition (SVD), whose low rank singular vectors constitute a compact representation. However, its factorized matrices contain negative values, which lacks a nice interpretation for co-clustering on documents or facial images. Therefore, we take advantage of a popularized tool, namely nonnegative matrix tri-factorization \cite{ding2006onmtf}, as the foundation of our approach. This method imposes the nonnegative constraints on the decomposed matrices, which leads to a parts-based representation \cite{lee1999nmf}. On the other hand, previous studies have shown human generated data (\textit{e.g.}, documents) are usually drawn from a probability distribution that has support on or near to a submanifold of the ambient Euclidean space \cite{cai2011lccf,cai2011gnmf,cai2008nmf-manifold}, and manifold learning has favorable applications in wide areas \cite{belkin2006manifold}. As a result, some researchers strive to consider manifold geometrical structure in co-clustering by dual graph regularization \cite{gu2009coclu-graph,wang2011snmtf,zhang2012ldcc}. However, it is a nontrivial and challenging task to seek the intrinsic manifolds of different types of data objects (\textit{e.g.}, samples and features) for graph based co-clustering. To address this issue, inspired from the work in \cite{geng2009emr}, we propose to approximate the optimal manifold by using a convex combination of some pre-given candidate manifolds, thus developing a novel method called \emph{Relational Multi-manifold Co-clustering} (RMC) to improve the clustering performance via manifold ensemble learning.

Our approach employs the symmetric nonnegative matrix tri-factorization to decompose the relational data matrix into three matrices for co-clustering. We consider the inter-type relation through the relational data matrix and the intra-type information through the affinity matrices constructed on both the sample and feature spaces \cite{wang2011snmtf, wang2011osntf}. In manifold ensemble learning, we assume that the intrinsic manifold of the sample or feature space lies in a convex hull of a group of pre-defined candidate manifolds \cite{geng2009emr}. It is strongly desirable to approximate the intrinsic manifolds of both the sample and feature spaces as much as possible, \textit{i.e.}, to learn an appropriate convex combination of a set of available manifolds. To optimize the objective function, the multiplicative update rules are adopted for the factorized matrices in an alternating manner. In this work, we exploit the \textit{entropic mirror descent algorithm} and the \textit{coordinate descent algorithm} to automatically learn the appropriate convex combination of candidate manifolds. To explore the performance of our method, we conducted extensive experiments on documents (\textit{e.g.}, webpages), images (\textit{e.g.}, handwritten digits and faces) and gene expression data. Experimental results suggest the efficacy of the proposed algorithm.

The remainder of this paper is organized as follows. In Section \ref{sec:relatedwork}, we briefly review the related works. Section \ref{sec:RMC} introduces our approach as well as two manifold coefficients optimization algorithms. Experimental results on several data sets from different domains are reported in Section \ref{sec:experiment} involving parameter selection and results analysis. Finally, we provide some concluding remarks in Section \ref{sec:conclusion}.

\section{Related Works}
\label{sec:relatedwork}
In this section, we primarily review some related works to our co-clustering method. In the last decade, co-clustering has become a hotspot topic and received lots of attention from the multidisciplinary communities because of its promising applications to many practical problems, \textit{e.g.}, collaborative filtering \cite{chen2009cf-onmtf,george2005cf-coclu}, microarray data analysis \cite{cho2004coclu-gene, hanisch2002coclu-bio}.

From the viewpoint of graph partitioning, two bipartite spectral graph partition methods were proposed in \cite{dhillon2001coclu-bipartite, zha2001bipartite} to co-cluster documents and words by finding minimum cut vertex partitions in a bipartite graph. More recently, authors in \cite{rege2006coclu-bipartie} put forward the isoperimetric co-clustering algorithm to partition the document-word bipartite graph, which minimizes the ratio of the perimeter of the bipartite graph partition and the partition area under a well-defined graph-theoretic area. For the information theory based co-clustering, samples and features are regarded as the instances of two random variables \cite{zhang2012ldcc}, \textit{e.g.}, an information bottleneck method was implemented in \cite{slonim2000mutualinfo} to cluster documents by using word clusters. Besides, an information-theoretic co-clustering algorithm specially designed for contingency table was introduced in \cite{dhillon2003coclu-info} and a more general co-clustering framework based on Bregman divergence was given in \cite{banerjee2004coclu-bregman}.

Matrix factorization based co-clustering techniques have been widely studied. At the very beginning, nonnegative matrix factorization was proposed to learn a parts-based representation, and it approximates the original data matrix by the product of two decomposed nonnegative matrices \cite{lee1999nmf,zhi2011gsnmf}. NMF has the intuitive interpretation for its results, but it focuses on the unilateral clustering and neglects the duality between rows and columns of a matrix. Motivated by this, the block value decomposition (BVD) was presented for co-clustering dyadic data \cite{long2005coclu-bvd}. It factorizes the data matrix into three components, \textit{i.e.}, the row and column coefficient matrices and the block value matrix. As an extension of NMF, orthogonal nonnegative matrix tri-factorization (ONMTF) was studied in \cite{ding2006onmtf}, which emphasizes the role of bi-orthogonality in three-factor NMF. Thanks to the successful applications of manifold learning in recent years \cite{belkin2006manifold,cai2009lpnmf,cai2009dyadic}, some researchers consider the local geometrical structure in matrix factorization based co-clustering. For example, a dual regularized co-clustering (DRCC) method based on semi-nonnegative matrix tri-factorization was proposed in \cite{gu2009coclu-graph}, which explores the geometrical structure of both data manifold and feature manifold. To reduce the computational complexity of DRCC, a fast nonnegative matrix tri-factorization approach was shown in \cite{wang2011fast-nmtf}, which constrains two factorized nonnegative matrices to be cluster indicator matrices. Moreover, a symmetric nonnegative matrix tri-factorization (SNMTF) framework was developed to cluster multi-type relational data \cite{wang2011snmtf}, which incorporates the intra-type information through manifold regularization. Based on this work, the authors further presented a fast nonnegative matrix tri-factorization approach to deal with large-scale data \cite{wang2011osntf}.

However, these graph regularized methods have a common shortcoming that the manifold used for co-clustering might not be the true intrinsic manifold, and it will even deviates far from the desired in an adverse situation. It is a nontrivial task to seek the intrinsic manifold in reality. To alleviate this difficulty, in light of \cite{geng2009emr}, we assume the intrinsic manifold of the sample or feature distribution lies in a convex hull of some candidate manifolds, and hope to maximally approximate the true intrinsic manifold using the convex combination of them. This way, the local geometrical structure can be better preserved in both the sample and feature spaces for co-clustering.

\section{Relational Multi-manifold Co-clustering}
\label{sec:RMC}
In this part, we introduce our RMC approach. We briefly review the symmetric NMTF based co-clustering of the multi-type relational data and describe the manifold ensemble learning, from which we arrive at our objective. The multiplicative update rules of the decomposed matrices as well as manifold coefficients optimization algorithms are provided. We begin with the problem formulation below.

\subsection{Problem Formulation}
The general problem setting is to co-cluster multi-type relational data. Given a $K$-type relational data set $\mathcal{X}=\{\mathcal{X}_1, \mathcal{X}_2, ..., \mathcal{X}_K\}$, where each $\mathcal{X}_k$ represents the data objects of the $k$-th type, we define an inter-type relational matrix $\bm{R}$ with the sub-matrix $\bm{R}_{ij}\in\mathbb{R}^{n_i\times n_j}$, $i\ne j$, which reflects the inter-type relationship between the $i$-th type and the $j$-th type data objects. To model the intra-type structure information for each data type, we define an intra-type relational matrix $\bm{W}$ consisting of a set of affinity matrices $\bm{W}_k\in \mathbb{R}^{n_k\times n_k}$ encoded on the data, which indicates the intra-type relation of components within the $k$-th type data. Both of the inter-type and intra-type relationship among the different data objects are to be fully used for co-clustering.

However, two-type relational data are omnipresent and frequently used in many real-world applications, \textit{e.g.}, simultaneously clustering documents and words, collaborative filtering in a recommendation system, gene expression data analysis under different experimental conditions. Hence, in this study we focus on the case $K=2$, \textit{i.e.}, we employ both the sample type and the feature type data objects for co-clustering. The concerned matrices can be respectively formulated as
\begin{displaymath}
\bm{R} = \left[ \begin{array}{cc}
\bm{0}^{n_1\times n_1}     & \bm{R}_{12}^{n_1\times n_2} \\
\bm{R}_{21}^{n_2\times n_1} & \bm{0}^{n_2\times n_2}
\end{array} \right],
\end{displaymath}
\begin{displaymath}
\bm{W} = \left[ \begin{array}{cc}
\bm{W}_{1}^{n_1\times n_1} &\bm{0}^{n_1\times n_2} \\
\bm{0}^{n_2\times n_1}    &\bm{W}_{2}^{n_2\times n_2}
\end{array} \right],
\end{displaymath}
where $\bm{0}$ is a matrix with all zero entries of different sizes, the superscripts denote the matrix sizes. Here, $\bm{R}_{12}$ and $\bm{R}_{21}$ represent the feature and sample matrix respectively, they satisfy the condition $\bm{R}_{12}=\bm{R}_{21}^T$. Each column of $\bm{R}_{21}$ or $\bm{R}_{12}$ denotes a feature or sample vector.

\subsection{Symmetric NMTF}
Symmetric nonnegative matrix tri-factorization (SNMTF) employs both the inter-type relationship and the intra-type information of multi-type relational data \cite{wang2011snmtf}. Similar to the previous works, \textit{i.e.}, Long's BVD \cite{long2005coclu-bvd}, Ding's ONMTF \cite{ding2006onmtf} and Gu's DRCC \cite{gu2009coclu-graph}, SNMTF shares an important property that they all decompose the data matrix into three low-rank matrices, \textit{i.e.}, ($K=2$)
\begin{equation}
\bm{R}_{12} \approx \bm{G}_1\bm{S}_{12}\bm{G}_2^T,
\end{equation}
where $\bm{G}_1\in \mathbb{R}^{n_1\times c_1}$ and $\bm{G}_2\in \mathbb{R}^{n_2\times c_2}$ are the cluster indicator matrices for $\mathcal{X}_1$ and $\mathcal{X}_2$, respectively, $c_1\ll n_1$, $c_2\ll n_2$. The middle matrix $\bm{S}_{12}\in \mathbb{R}^{c_1\times c_2}$ can be treated as a compact representation of $\bm{R}_{12}$ \cite{long2005coclu-bvd}, which absorbs the different scales of other matrices \cite{ding2006onmtf}.

Note that BVD and ONMTF impose nonnegative constraints on all three matrices $\bm{G}_1$, $\bm{G}_2$, $\bm{S}$, while DRCC and SNMTF both relax the nonnegative constraint on the matrix $\bm{S}$, thereby allowing negative entries. Different from BVD, ONMTF emphasizes the bi-orthogonality on $\bm{G}_1$, $\bm{G}_2$. Compared to ONMTF, both DRCC and SNMTF consider the local geometrical structure by incorporating the Laplacian regularization into the objective. Based on DRCC, SNMTF employs the symmetric matrix by virtue of the idea in \cite{long2006sc-multirelation,wang2008ssc-nmf} to simultaneously cluster multi-type relational data, and its objective can be shown by
\begin{equation} \label{obj:SNMTF}
\min_{\bm{G},\bm{S}} \|\bm{R}-\bm{GSG}^T\|_F^2 + 2\lambda \tr(\bm{G}^T\bm{LG}), ~~ s.t. ~ \bm{G}\succeq 0,
\end{equation}
where $\|\cdot\|_F$ denotes the Frobenius norm of a matrix, $\bm{L}=\bm{D}-\bm{W}$ is the graph Laplacian \cite{belkin2006manifold}, $\bm{D}$ is the diagonal degree matrix with $\bm{D}_{k_{(ii)}}=\sum_j \bm{W}_{k_{(ij)}}$, and $\bm{L}_k = \bm{D}_{k}- \bm{W}_k$.

The relational matrices $\bm{G}$ and $\bm{S}$ are designed similarly to $\bm{W}$ and $\bm{R}$, respectively, \textit{i.e.},
\begin{displaymath}
\bm{G} = \left[ \begin{array}{cc}
\bm{G}_{1}^{n_1\times c_1} &\bm{0}^{n_1\times c_2} \\
\bm{0}^{n_2\times c_1}    &\bm{G}_{2}^{n_2\times c_2}
\end{array} \right],
\end{displaymath}
\begin{displaymath}
\bm{S} = \left[ \begin{array}{cc}
\bm{0}^{c_1\times c_1}     & \bm{S}_{12}^{c_1\times c_2} \\
\bm{S}_{21}^{c_2\times c_1} & \bm{0}^{c_2\times c_2}
\end{array} \right].
\end{displaymath}

\subsection{Manifold Ensemble Learning}
As can be observed from the existing graph based co-clustering methods, the intrinsic manifold plays a crucial role in preserving the local geometrical structure in the data space. However, it is a nontrivial task to discover an appropriate intrinsic manifold in reality. Therefore, automatic and data-driven manifold approximation is said to be invaluable for manifold regularization based co-clustering. In this work, we adopt a novel learning paradigm named \textit{manifold ensemble learning} to maximally approximate the true intrinsic manifold. This idea is inspired from the work in \cite{geng2009emr}, which combines the automatic intrinsic manifold approximation and semi-supervised classification.

Our assumption is that a series of initial guesses of graph Laplacian are available and the intrinsic manifold of the sample or feature space lies in the convex hull of these pre-given candidate manifolds. In some sense, this assumption constrains the search space of possible manifolds, since the optimal graph Laplacian is an discrete approximation to the intrinsic manifold \cite{geng2009emr}, \textit{i.e.},
\begin{equation} \label{eq:manifoldensemble}
\bm{L} = \sum_{i=1}^q \mu_i \tilde{\bm{L}}_i, ~ s.t.~\sum_{i=1}^q\mu_i=1, \mu_i\ge 0,
\end{equation}
where a set of candidate graph Laplacians $C=\{\bm{L}_1, \ldots, \bm{L}_q\}$ is defined and the convex hull of this set is denoted by
\begin{displaymath}
\textbf{conv}C=\{\sum_i^q \mu_ix_i| \sum_i \mu_i=1, x_i\in C, x_i\ge0\}.
\end{displaymath}
Here we use $\tilde{\bm{L}}_i$ of the $i$-th candidate manifold to discriminate it from $\bm{L}_k$ of the $k$-th type data. Now, we have $\bm{L}\in \textbf{conv}C$, which is also a graph Laplacian.

\subsection{Objective Function}
Based on the above analysis, it is natural for us to take advantage of the manifold ensemble learning to approximate the intrinsic manifold in the sample and feature space, respectively. In concrete, we incorporate this idea into the symmetric nonnegative matrix tri-factorization framework and propose a novel co-clustering approach named \textit{Relational Multi-manifold Co-clustering} (RMC).

Now, it is easy to arrive at the objective function, \textit{i.e.},
\begin{equation} \label{obj:RMC}
\begin{aligned}
\min_{\bm{G},\bm{S},\bm{\mu}}
&\|\bm{R}-\bm{GSG}^T\|_F^2 + \alpha \tr [\bm{G}^T(\sum_{i=1}^q\mu_i\tilde{\bm{L}}_i)\bm{G}] + \beta\|\bm{\mu}\|_2^2,\\
 s.t. & ~ \sum_{i=1}^q\mu_i=1, \bm{\mu}\succeq 0, \bm{G}\succeq 0,
\end{aligned}
\end{equation}
where $\alpha>0, \beta>0$, the tradeoff parameter $\alpha$ is used to govern the contribution of the ensemble manifold regularization to the objective, the $l_2$-norm of $\bm{\mu}$ is employed to avoid the coefficient parameter over-fitting to only one manifold and the factor $\beta$ acts as an over-fitting tolerance parameter for the manifold coefficients. Similar to \cite{gu2009coclu-graph}, we do $l_2$ normalization on columns of $\bm{G}$ and compensate its norm to $\bm{S}$. The components of $\bm{G}$, \textit{i.e.}, $\bm{G}_1$ and $\bm{G}_2$ represent the partition matrices of the feature matrix $\bm{R}_{21}$ and the sample matrix $\bm{R}_{12}$, respectively. We typically use the partition matrices to derive the co-clustering results.
\subsection{Optimization}
In this section, we explore how to optimize the objective in Eq.~(\ref{obj:RMC}). It can be readily found that the objective function is non-convex in $\bm{G},\bm{S},\bm{\mu}$ jointly, but it is convex in them respectively. So it is unrealistic to find the global minimum since no closed-form solution can be obtained. We present an alternating scheme to optimize the objective as most of the previous works do \cite{cai2011gnmf, gu2009coclu-graph, long2005coclu-bvd, wang2011snmtf}. However, different from the existing schemes, it is more challenging to optimize our objective since it has a critical manifold coefficient vector to be solved. Details are shown below.

\subsubsection{Computation of $\bm{S}$}
When fixing $\bm{G}$ and $\bm{\mu}$, the objective becomes minimizing $J_S=\|\bm{R}-\bm{GSG}^T\|_F^2$. Taking its derivative to $\bm{S}$, \textit{i.e.},
\begin{displaymath}
\frac{\partial J_S}{\partial \bm{S}} = -2\bm{G}^T\bm{RG} + 2\bm{G}^T\bm{GSG}^T\bm{G},
\end{displaymath}
and setting it to 0, then we have the update rule
\begin{equation} \label{updateS}
\bm{S} = (\bm{G}^T\bm{G})^{-1}\bm{G}^T\bm{RG}(\bm{G}^T\bm{G})^{-1}.
\end{equation}

\subsubsection{Computation of $\bm{G}$}
When fixing $\bm{S}$ and $\bm{\mu}$, the objective with respect to $\bm{G}$ reduces to minimizing
\begin{displaymath}
J_G = \|\bm{R}-\bm{GSG}^T\|_F^2 + \alpha \tr(\bm{G}^T\bm{L}\bm{G}), ~s.t.~\bm{G}\succeq 0.
\end{displaymath}

To solve this constrained optimization problem, we introduce the Lagrangian multiplier matrix $\bm{\Lambda}$ and its Lagrangian function is formulated as
\begin{equation}
L(\bm{G}) = \|\bm{R}-\bm{GSG}^T\|_F^2 + \alpha \tr(\bm{G}^T\bm{L}\bm{G}) + \tr(\bm{\Lambda}\bm{G}^T).
\end{equation}
Requiring its derivative to $\bm{G}$ be 0, we obtain
\begin{displaymath}
\bm{\Lambda} = 4\alpha\bm{LG}-4\bm{A}+4\bm{GB},
\end{displaymath}
where $\bm{A}=\bm{RGS}^T$, $\bm{B}=\bm{S}^T\bm{G}^T\bm{GS}$.

Since the Karush-Kuhn-Tucker (KKT) complementary condition \cite{boyd2004convex} for the nonnegativity of $\bm{G}_{ij}$ gives $\bm{\Lambda}_{ij}\bm{G}_{ij}=0$, we have
\begin{equation} \label{eq:KKT}
(\alpha\bm{LG}- \bm{A} + \bm{GB})_{ij}\bm{G}_{ij} = 0.
\end{equation}

Similar to \cite{ding2010convex}, we define $\bm{L} = \bm{L}^+ - \bm{L}^-$, $\bm{A} = \bm{A}^+ - \bm{A}^-$, $\bm{B} = \bm{B}^+ - \bm{B}^-$, where
\begin{gather}
\bm{L}_{ij}^+ = \frac{(|\bm{L}_{ij}| + \bm{L}_{ij})}{2}, \notag ~~
\bm{L}_{ij}^- = \frac{(|\bm{L}_{ij}| - \bm{L}_{ij})}{2}. \notag
\end{gather}
We substitute the decomposed positive and negative parts into Eq.~(\ref{eq:KKT}), which leads to the update rule
\begin{equation} \label{updateG}
\bm{G}_{ij}\leftarrow \bm{G}_{ij}\left[
\frac{(\alpha\bm{L}^-\bm{G}+ \bm{A}^+ + \bm{GB}^-)_{ij}}{(\alpha\bm{L}^+\bm{G}+ \bm{A}^- + \bm{GB}^+)_{ij}}
\right]^{\frac{1}{2}}.
\end{equation}

\subsubsection{Computation of $\bm{\mu}$}
When fixing $\bm{G}$ and $\bm{S}$, the objective is simplified to
\begin{equation}
\begin{aligned}
\min_{\bm{\mu}} &~f(\bm{\mu})=\sum_{i=1}^q\mu_i s_i + \beta\|\bm{\mu}\|_2^2, \\
~s.t.& ~\sum_{i=1}^q \mu_i=1, \bm{\mu}\succeq 0,
\end{aligned}
\end{equation}
where $s_i=\tr(\bm{G}^T \tilde{\bm{L}}_i\bm{G})$. It is easy to see that if $\beta=0$, then the trivial solution will be
\begin{displaymath}
\mu_i= \left\{ \begin{array}{ll}
1, & \textrm{if $s_i=\min_{k=1,\ldots,n} s_k$},\\
0, & \textrm{otherwise}. \end{array}\right.
\end{displaymath}
This is extremely sparse and undesirable to learn a composite manifold. On the other hand, if $\beta \rightarrow \infty$, all the candidate manifolds will receive identical weights, which is also unexpected. Therefore, it is essential to assign a reasonable value for the parameter $\beta$.
%
%
\begin{algorithm}[t]
 \caption{Entropic Mirror Descent Algorithm (EMDA)}
 \label{alg:EMDA}
 \begin{algorithmic}[1]
  \REQUIRE~~\\
  The Lipschitz constant $L_f$, \\
  The tradeoff parameter $\beta$, \\
  The ensemble manifold $\bm{s}$.
  \ENSURE~~\\
  The manifold coefficient vector $\bm{\mu}$.   

  \PROCEDURE
  \STATE Initialize $\mu_i$ with the identical weight $1/q$, where $q$ is the number of candidate manifolds.
  \FOR{$i = 1$ to $q$}

  \REPEAT
  \STATE ~~Compute $t_m = \sqrt{\frac{2\ln q}{m L_f^2}}$, where $m$ is the $m$-th iteration.
  \STATE ~~Update each $\mu_i$ according to
   \begin{displaymath}
         ~~~~~~\mu_i^{m+1} \leftarrow
         \frac{\mu_i^m \exp[-t_m f'(\mu_i^m)]}{\sum_{i=1}^q\mu_i^m \exp[-t_m f'(\mu_i^m)]},
   \end{displaymath}
         ~~where $ f'(\mu_i^m) = 2\beta\mu_i^m + s_i.$
  \UNTIL convergence

  \ENDFOR
 \end{algorithmic}
\end{algorithm}
%
\begin{algorithm}[t]
\caption{Coordinate Descent Algorithm (CDA)}
 \label{alg:CDA}
 \begin{algorithmic}[1]
  \REQUIRE~~\\
  The tradeoff parameter $\beta$, \\
  The ensemble manifold $\bm{s}$. 
  \ENSURE~~\\
  The manifold coefficient vector $\bm{\mu}$.   

  \PROCEDURE
  \STATE Initialize $\mu_i$ with the identical weight $1/q$, where $q$ is the number of candidate manifolds.
  \FOR{$i = 1$ to $q$}
  \FOR{$j = 1$ to $q$ ($j\ne i$)}

  \REPEAT
  \IF {$2\beta(\mu_i+\mu_j) + (s_j-s_i)\le 0$}
  \STATE $\mu_i^* = 0, ~\mu_j^*=\mu_i+\mu_j.$
  \ELSE
     \IF {$2\beta(\mu_i+\mu_j) + (s_i-s_j)\le 0$}
     \STATE $\mu_j^* = 0, ~\mu_i^*=\mu_i+\mu_j'.$
     \ELSE
     \STATE $\mu_i^* =\frac{2\beta(\mu_i+\mu_j)+(s_j-s_i)}{4\beta},$
     \STATE $\mu_j^*=\mu_i+\mu_j-\mu_i^*.$
     \ENDIF
  \ENDIF

  \UNTIL convergence

  \ENDFOR
  \ENDFOR
 \end{algorithmic}
\end{algorithm}
\begin{algorithm}[t]
\caption{Relational Multi-manifold Co-clustering (RMC)}
 \label{alg:RMC}
 \begin{algorithmic}[1]
  \REQUIRE~~\\
  The relational data matrices $\bm{R}$ and $\bm{W}$, \\
  The number of co-clusters $c_1$ and $c_2$, \\
  The tradeoff parameters $\alpha$ and $\beta$, \\
  The convergence rate $\epsilon = 10^{-5}$.   
  \ENSURE~~\\
  The partition matrix $\bm{G}$.   

  \PROCEDURE
  \STATE Initialize $\bm{G}$ using k-means.

  \REPEAT
  \STATE \label{rmc1}Compute $\bm{S} = (\bm{G}^T\bm{G})^{-1}\bm{G}^T\bm{RG}(\bm{G}^T\bm{G})^{-1}.$
  \STATE Learn the manifold coefficient vector $\bm{\mu}$ using EMDA in Algorithm~\ref{alg:EMDA} or CDA in Algorithm~\ref{alg:CDA}.
  \STATE \label{rmc2} Update the matrix $\bm{G}$ according to
   \begin{displaymath}
   \bm{G}_{ij}\leftarrow \bm{G}_{ij}\left[
   \frac{(\alpha\bm{L}^-\bm{G}+ \bm{A}^+ + \bm{GB}^-)_{ij}}{(\alpha\bm{L}^+\bm{G}+ \bm{A}^- + \bm{GB}^+)_{ij}}
   \right]^{\frac{1}{2}}.
   \end{displaymath}
  \UNTIL convergence

 \end{algorithmic}
\end{algorithm}

To optimize this problem, there are generally three possible ways. First, it can be solved by the generic \textit{Quadratic Programming} (QP) method (e.g, CVX)\cite{boyd2004convex}, but this solver is often time-consuming for larger size problem and shows slow convergence. Second, it is actually an exactly well-defined problem, \textit{i.e.}, the convex minimization over the unit simplex, which can be elegantly solved by the \textit{Entropic Mirror Descent Algorithm} (EMDA) with a global efficiency estimate proven to be mildly dependent on the problem size \cite{beck2003mirror}, as shown in Algorithm~\ref{alg:EMDA}. Third, we can adopt the \textit{Coordinate Descent Algorithm} (CDA) just like using sequential minimal optimization for support vector machines \cite{platt1999smo}, which selects two variables to update in each iteration while keeping the others fixed, as shown in Algorithm~\ref{alg:CDA}. In this work, we put emphasis on EMDA and CDA, which are employed to learn appropriate manifold coefficients. The brief descriptions about them are given below.

\textbf{Entropic Mirror Descent Algorithm}. It can be viewed as a nonlinear projected-subgradient type method, derived from using a general distance-like function instead of the usual Euclidean squared distance \cite{beck2003mirror}. It has been shown that EMDA owns the natural advantage to solve this convex problem over the unit simplex $\triangle=\{\bm{\mu}\in\mathbb{R}^q: \sum_{i=1}^q \mu_i=1, \bm{\mu}\succeq 0\}$. To apply this algorithm, the objective function $f$ should be a convex Lipschitz continuous function with Lipschitz constant $L_f$ with respect to a fixed given norm. In our approach, we derive this Lipschitz constant from $\|\nabla f(\bm{\mu})\|_1\le2\beta+\|\bm{s}\|_1=L_f$, where $\bm{s}=\{s_1,\ldots,s_q\}$. Here, we use $\|\cdot\|_1$ norm as suggested in \cite{beck2003mirror}.

\textbf{Coordinate Descent Algorithm}. In this method, a pair of variables are selected to joint the update process while holding the others fixed in each iteration. It is true that the summation of each pair of elements will keep still after the previous iteration due to the constraint $\sum_{i=1}^q \mu_i=1$. All pairs of elements in the coefficient vector $\bm{\mu}$ will be iteratively scanned. Compared to EMDA, CDA costs more time since it has to traverse over all the element pairs while EMDA only goes over the elements in sequence.
\subsection{Our RMC Approach}
In summary, we present the primary procedures of the proposed \textit{Relational Multi-manifold Co-clustering} (RMC) approach in Algorithm~\ref{alg:RMC}.

As can be seen, we will finally obtain a partition matrix $\bm{G}$ used for co-clustering the samples and features simultaneously. Since the intrinsic manifold is maximally approximated via manifold ensemble learning, the local geometrical structure of the sample or feature space can be better respected, thus achieving more promising co-clustering results. In particular, we define the RMC approach using EMDA to learn the manifold coefficient as RMC-E for short, and using CDA as RMC-C for short.

Note that here we omit the convergence proof of the multiplicative update rules, since our method essentially follows the similar fashion of many existing co-clustering algorithms, \textit{e.g.,} DRCC \cite{gu2009coclu-graph}, SNMTF \cite{wang2011snmtf}. To probe it deeply, we refer the readers to these literatures for details. Besides, for matrix factorization based co-clustering methods, there is an important issue to be considered, \textit{i.e.,}, computational complexity. Luckily, there have been some research works focusing on this problem in \cite{wang2011osntf, wang2011fast-nmtf}. Therefore, to make our algorithm scalable, it is an ideal choice to adopt those speed-up strategies.
\section{Experiments}
\label{sec:experiment}
In this section, we aim to investigate the clustering performance of the proposed method on a broad range of collected data sets. A number of interesting experiments were carried out to demonstrate the effectiveness of our approach for document, image and gene expression data clustering, respectively. We first give brief descriptions about the data sets and evaluation criteria. Then, parameter settings for all the compared algorithms are presented and the corresponding results are reported. Finally, we explore the parameter selection and the distributions of the manifold coefficients.

\subsection{Data Corpora}
We conduct the performance evaluations on several diverse data collections, \textit{i.e.}, two text corpora, three image databases and three gene expression data. Their important statistics are summarized in Table~\ref{table:dataset} and the brief descriptions about them are shown below.
\begin{table}[t]
\centering
\caption{Statistics of the data sets}
\label{table:dataset}
\begin{tabular}{lcrrc} \hline \hline
Data Sets  &Domain  &Samples    &Features     &Classes \\\hline
NGroups5   &text    &4,052      &1,200        &5  \\
RCV1-5     &text    &3,012      &1,200        &5  \\
COIL20     &image   &1,440      &1,024        &20  \\
AlphaDigit &image   &1,404      & 320         &36  \\
UMIST      &image   &575        &644          &20  \\
Leukemia2  &gene    &72         &5,551        &3  \\
LungCancer &gene    &203        &2,008        &5  \\
SRBCT      &gene    &83         &2,308        &4  \\
\hline\end{tabular}
\end{table}

\textbf{Text corpora}. NGroups5 is selected from the popular newsgroup data collection 20Newsgroups\footnote{http://people.csail.mit.edu/jrennie/20Newsgroups/}. We use a subset of the 20news-bydate version, which removed the duplicates and some headers. This subset contains five different topics, which refer to 4,052 documents. RCV1-5 is a subset chosen from the Reuters Corpus Volume I\footnote{http://www.daviddlewis.com/resources/testcollections/rcv1/} (RCV1) collection \cite{lewis2004rcv1}. We choose five topics contained in a smaller RCV1 database \cite{chen2011parallel-sc}, and this subset is associated with the \textit{`M141',`GCAT',`G151',`G158', `G159'} topics. For the two data sets, we preprocess them using a standard feature selection mechanism in \cite{slonim2000mutualinfo} for text data, \textit{i.e.}, we select 1,200 words with the highest contribution to the mutual information between the words and the documents \cite{chen2010nmf-semi-coclu}.

\textbf{Image databases}. The COIL20\footnote{http://www1.cs.columbia.edu/CAVE/software/softlib/coil-20.php} image library contains 20 different objects viewed from varying angles. Each object has 72 gray scale images with the size of 32$\times$32. The AlphaDigit\footnote{http://cs.nyu.edu/$\sim$roweis/data.html} is a handwritten image database that contains 1,404 binary images, covering 20$\times$16 digits of ``0" through ``9" and capitals ``A" through ``Z". UMIST\footnote{http://images.ee.umist.ac.uk/danny/database.html} is a face image database that contains 575 multi-view face images of 20 people, referring to a range of poses from profile to frontal views. Each image is rescaled to 28$\times$23 pixels.

\textbf{Gene expression data}. The three microarray gene expression data\footnote{http://www.gems-system.org/datasets/} Leukemia2, LungCancer and SRBCT are often used to do multicategory cancer diagnosis in bioinformatics \cite{statnikov2005gene-bio}. Leukemia2 and LungCancer are produced by oligonucleotide-based technology. SRBCT was obtained by using two-color cDNA platform with consecutive image analysis. Similar to \cite{liu2008alphagene}, we removed the genes which vary little across samples so as to reduce the computational complexity. Table~\ref{table:gene} shows the exact preprocessing strategy of the gene expression data. Details about each term can be referred to \cite{liu2008alphagene}. No operation is conducted on SRBCT since it has been originally preprocessed in \cite{khan2001srbct}.
%
\begin{table}[t]
\centering
\caption{Preprocessing strategy of the gene data}
\label{table:gene}
\begin{tabular}{lcccc} \hline \hline
Datasets   &Floor  &Ceiling  &Max/Min  &Max-Min  \\ \hline
Leukemia2  &100   &16000     &25       &500     \\
LungCancer &0     &16000     &5        &500     \\
SRBCT      &-     &-         &-        &-     \\
\hline\end{tabular}
\end{table}

Notice that all data sets used here are normalized to unit Euclidean length in prior. Besides, the entries of each input data matrix are nonnegative, thereby all compared algorithms can be applied to yield the clustering results.

\subsection{Performance Comparison}
To explore the clustering performance of the proposed algorithm, we compare it with some state-of-the-art approaches, which are clearly listed below.
\begin{itemize}
\item \textbf{KM}:~Conventional k-means method.
\item \textbf{NMF}:~Nonnegative matrix factorization \cite{lee1999nmf}.
\item \textbf{GNMF}:~Graph regularized nonnegative matrix factorization, which considers local geometrical structure by the sample graph regularization \cite{cai2011gnmf}.
\item \textbf{DRCC}:~Dual regularized co-clustering, which employs both the sample and feature graph regularization \cite{gu2009coclu-graph}.
\item \textbf{ONMTF}:~Orthogonal nonnegative matrix tri-factoriza-tion with the bi-orthogonality constraints imposed \cite{ding2006onmtf}.
\item \textbf{SNMTF}:~Symmetric nonnegative matrix tri-factoriza-tion, which utilizes NMTF to simultaneously cluster different types of data \cite{wang2011snmtf}.
\item \textbf{OSNTF}:~Orthogonal symmetric nonnegative matrix tri-factorization, which is designed for simultaneous clustering of multi-type relational data with the orthogonality constraint \cite{wang2011osntf}.
\item \textbf{RMC}:~Our relational multi-manifold co-clustering approach, which makes use of the entropic mirror descent algorithm (RMC-E) and the coordinate descent algorithm (RMC-C) respectively to learn a convex combination of a group of candidate manifolds.
\end{itemize}

Note that among the above compared methods, the original update rule of $\bm{G}$ in SNMTF \cite{wang2011snmtf} and OSNTF \cite{wang2011osntf} might appear negative elements due to the polarity uncertainty of the numerator $(\bm{RGS}+\lambda \bm{WG})$. In consequence, we follow \cite{gu2009coclu-graph} and use the similar update rule for the matrix $\bm{G}$ in SNMTF and OSNTF, so that the nonnegativity can be well guaranteed and the objective strictly reduces. For both SNMTF and OSNTF, the cluster membership of each sample is determined by the corresponding row vector of the cluster indicator matrix $\bm{G}$. For the remaining methods except KM, we perform clustering using k-means in the derived low-dimensional data space. In the experiments, we ran k-means 20 times with different randomly generated starting points and the result in terms of the minimized objective function was recorded. For all the evaluated approaches, we repeat the clustering twenty times and the average results over 20 test runs are reported.
%
\begin{table*}[t]
\caption{Clustering accuracy of different algorithms ($\%$)}
\centering
\label{table:AC}
\begin{tabular}[!ht]{lccccccccc} \hline \hline
Data Sets   &~~KM~~ &~NMF~  &GNMF~  &DRCC~  &ONMTF  &SNMTF  &OSNTF  &RMC-E            &RMC-C \\\hline
NGroups5   &30.70  &36.92  &44.03  &50.37  &38.15  &47.09  &45.56  &\textbf{51.99*}   &50.97~ \\
RCV1-5     &54.52  &55.12  &58.53  &61.81  &58.08  &60.70  &60.82  &\textbf{64.67*}   &62.37~ \\
COIL20     &63.35  &62.70  &73.13  &73.91  &60.87  &71.48  &70.42  &\textbf{76.81*}   &76.46~ \\
AlphaDigit &43.09  &39.67  &41.10  &45.41  &42.18  &45.29  &42.88  &\textbf{46.37~}   &45.53~ \\
UMIST      &40.60  &39.83  &53.91  &55.36  &41.80  &57.04  &54.78  &\textbf{60.52*}   &57.08~ \\
Leukemia2  &65.07  &64.03  &87.50  &88.04  &68.25  &88.40  &72.22  &\textbf{90.28*}   &\textbf{90.28*} \\
LungCancer &70.99  &64.31  &82.76  &83.45  &72.10  &84.36  &75.86  &\textbf{89.66*}   &\textbf{89.66*} \\
SRBCT      &48.73  &41.20  &53.01  &59.06  &49.34  &60.24  &51.81  &61.45~            &\textbf{63.06*} \\
\hline\end{tabular}

\centering
\caption{Normalized mutual information of different algorithms ($\%$)}
\label{table:NMI}
\begin{tabular}[!ht]{lccccccccc} \hline \hline
Data Sets   &~~KM~~ &~NMF~  &GNMF~  &DRCC~  &ONMTF  &SNMTF  &OSNTF  &RMC-E            &RMC-C \\\hline
NGroups5   &13.41  &16.62  &27.69  &36.11  &18.66  &35.02  &28.70  &\textbf{37.65*}   &36.30~ \\
RCV1-5     &44.06  &45.19  &48.78  &51.50  &46.25  &48.94  &48.64  &\textbf{52.02~}   &50.97~ \\
COIL20     &74.32  &72.68  &83.17  &85.84  &73.02  &84.21  &83.78  &\textbf{88.35*}   &86.48~ \\
AlphaDigit &58.40  &55.22  &56.01  &60.21  &57.44  &60.69  &56.44  &\textbf{62.28*}   &59.48~ \\
UMIST      &60.14  &58.73  &71.03  &72.30  &59.73  &73.72  &70.97  &\textbf{76.61*}   &73.97~ \\
Leukemia2  &53.65  &49.19  &65.81  &68.25  &54.11  &69.45  &49.43  &\textbf{71.45*}   &\textbf{71.45*} \\
LungCancer &54.53  &53.70  &63.46  &65.20  &55.69  &71.07  &54.71  &\textbf{73.97*}   &72.01~ \\
SRBCT      &25.80  &20.65  &31.67  &34.48  &23.61  &33.45  &24.46  &35.06~            &\textbf{37.43*} \\
\hline\end{tabular}
\end{table*}
\subsection{Evaluation Criteria}
We adopt two popular criteria to measure the clustering performance \cite{cai2005document-lpi,gu2009coclu-graph}, \textit{i.e.}, the \emph{accuracy} (AC) and the \emph{normalized mutual information} (NMI).

AC denotes the percentage of correct labels estimated by the clustering algorithm. Given a data point $\bm{x}_i$, let $a_i$ and $g_i$ be the estimated and true label respectively, then we have
\begin{equation}
 \text{AC}=\frac {\sum_{i=1}^n \delta(g_i, map(a_i))}{n},
\end{equation}
where $n$ is the total number of samples, $\delta(\cdot, \cdot)$ is an indicator function that equals one if the two entries are the same and equals zero otherwise. The permutation mapping function $map(a_i)$ maps each cluster label $a_i$ to the equivalent label of the data set.

NMI evaluates how closely the clustering algorithm is able to reconstruct the underlying label distribution in the data corpus. Let $C$ and $C'$ be the cluster sets from the ground truth and the clustering method respectively, then NMI between them is defined as
\begin{equation}
 \textrm{NMI}(C,C')=\frac{\sum_{c_i\in C, c_j'\in C'} p(c_i,c_j') \cdot \log_2 \frac {p(c_i,c_j')}{p(c_i)\cdot p(c_j')}}{\max(H(C),H(C'))},
\end{equation}
where the probabilities $p(c_i)$, $p(c_j')$ indicate to what extent a sample belongs to the clusters $c_i$ and $c_j'$ respectively. $p(c_i, c_j')$ is the joint probability that the selected sample belongs to $c_i$ and $c_j'$ simultaneously. $H(C)$ and $H(C')$ are the entropies of $C$ and $C'$. $\textrm{NMI}(C,C')$ takes values between zero and one. The larger the NMI value is, the better clustering performance the algorithm will achieve.
%
%
%
\begin{figure*}[!ht]
\centering
\subfigure[NGroups5]{\includegraphics[width=0.23\textwidth]{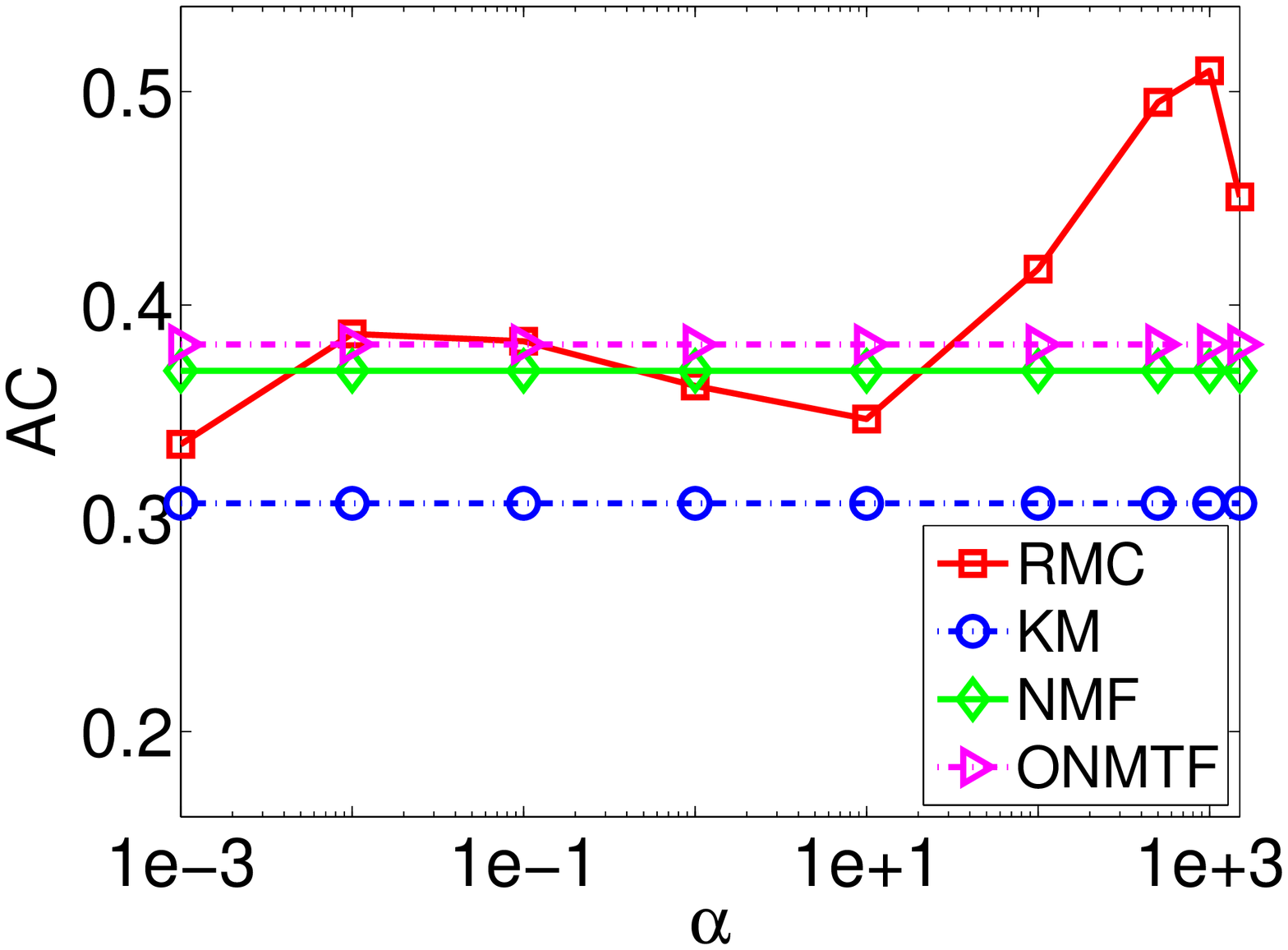}}
\subfigure[RCV1-5]{\includegraphics[width=0.23\textwidth]{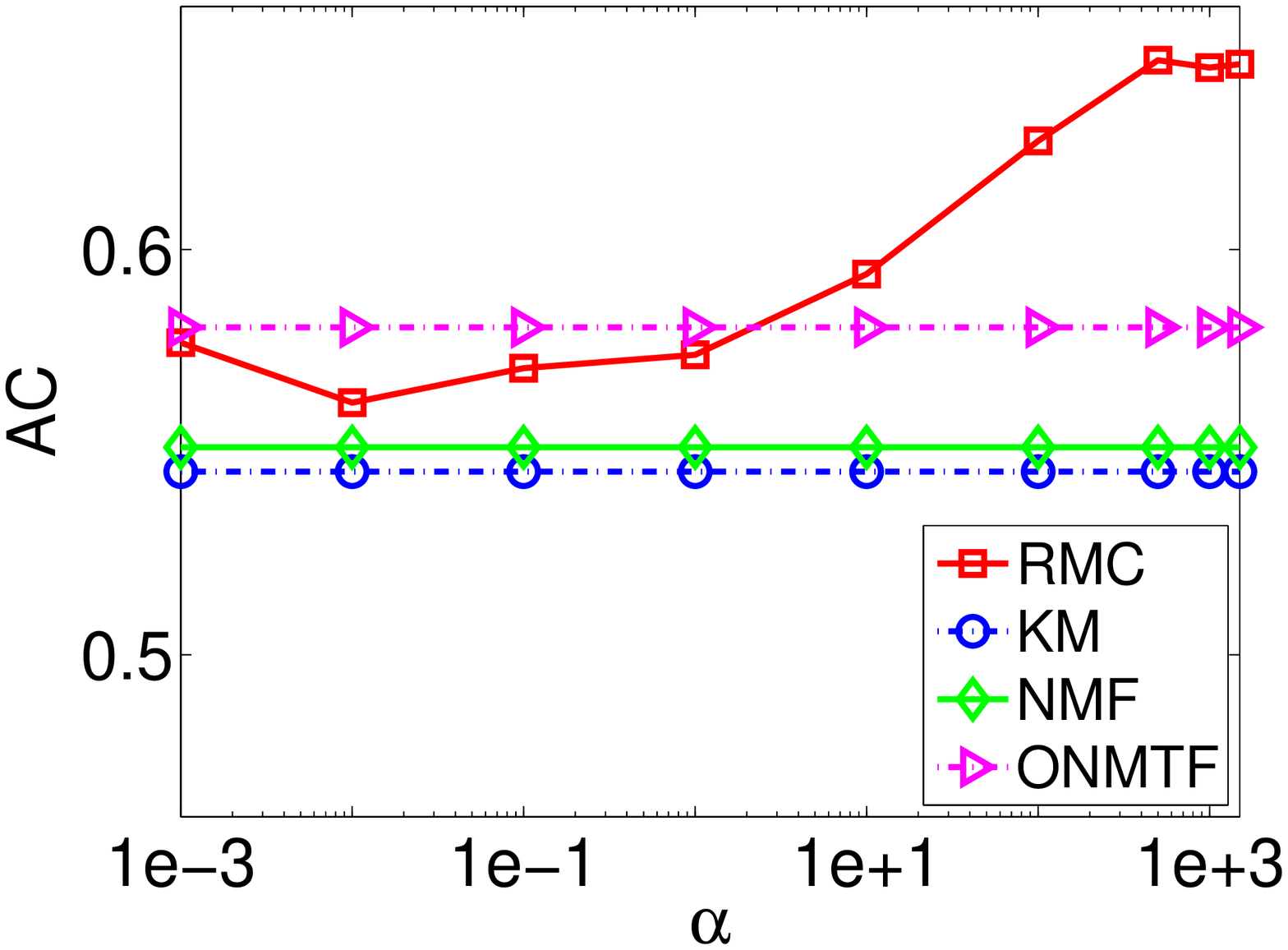}}
\subfigure[COIL20]{\includegraphics[width=0.23\textwidth]{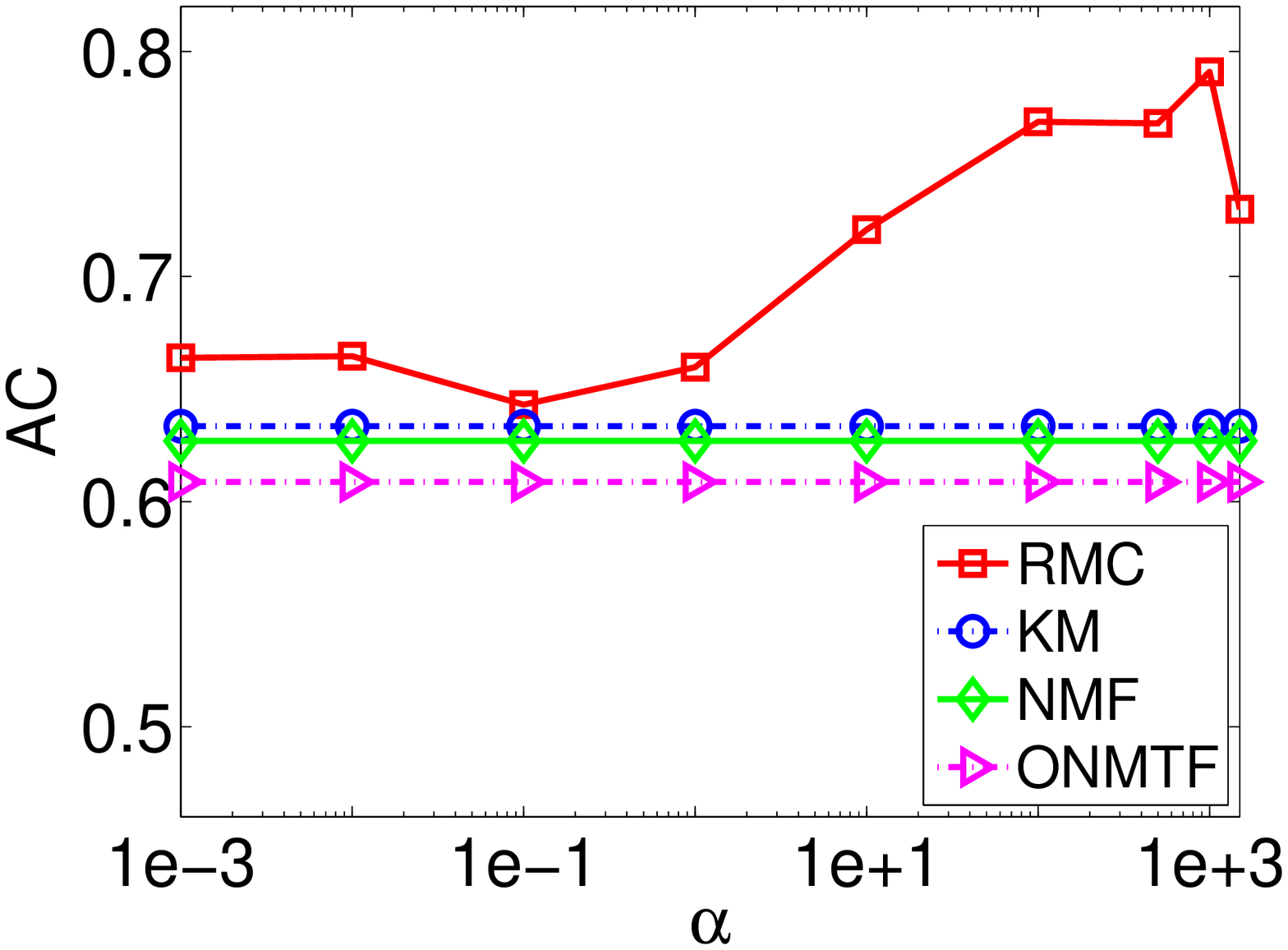}}
\subfigure[AlphaDigit]{\includegraphics[width=0.23\textwidth]{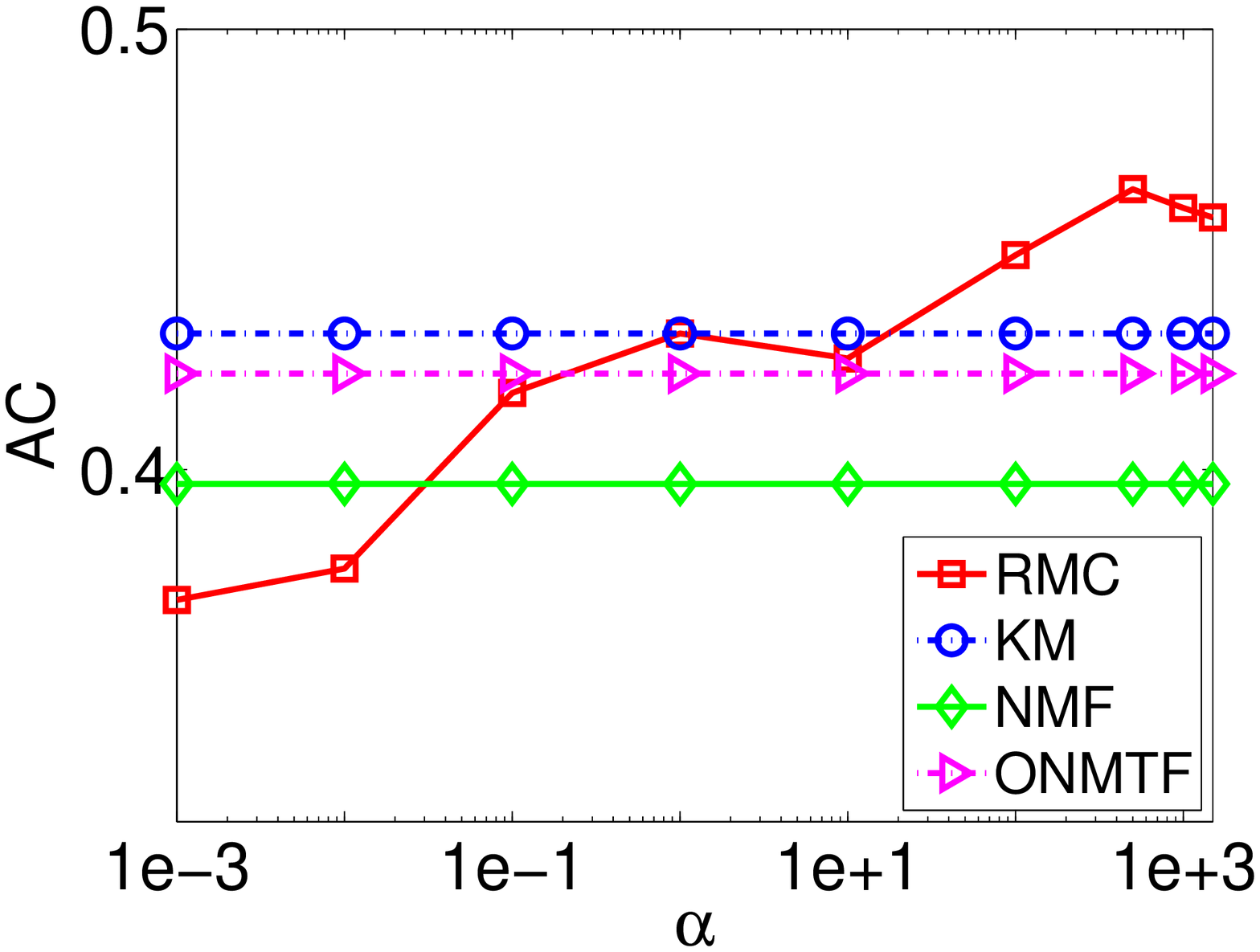}}
\subfigure[UMIST]{\includegraphics[width=0.23\textwidth]{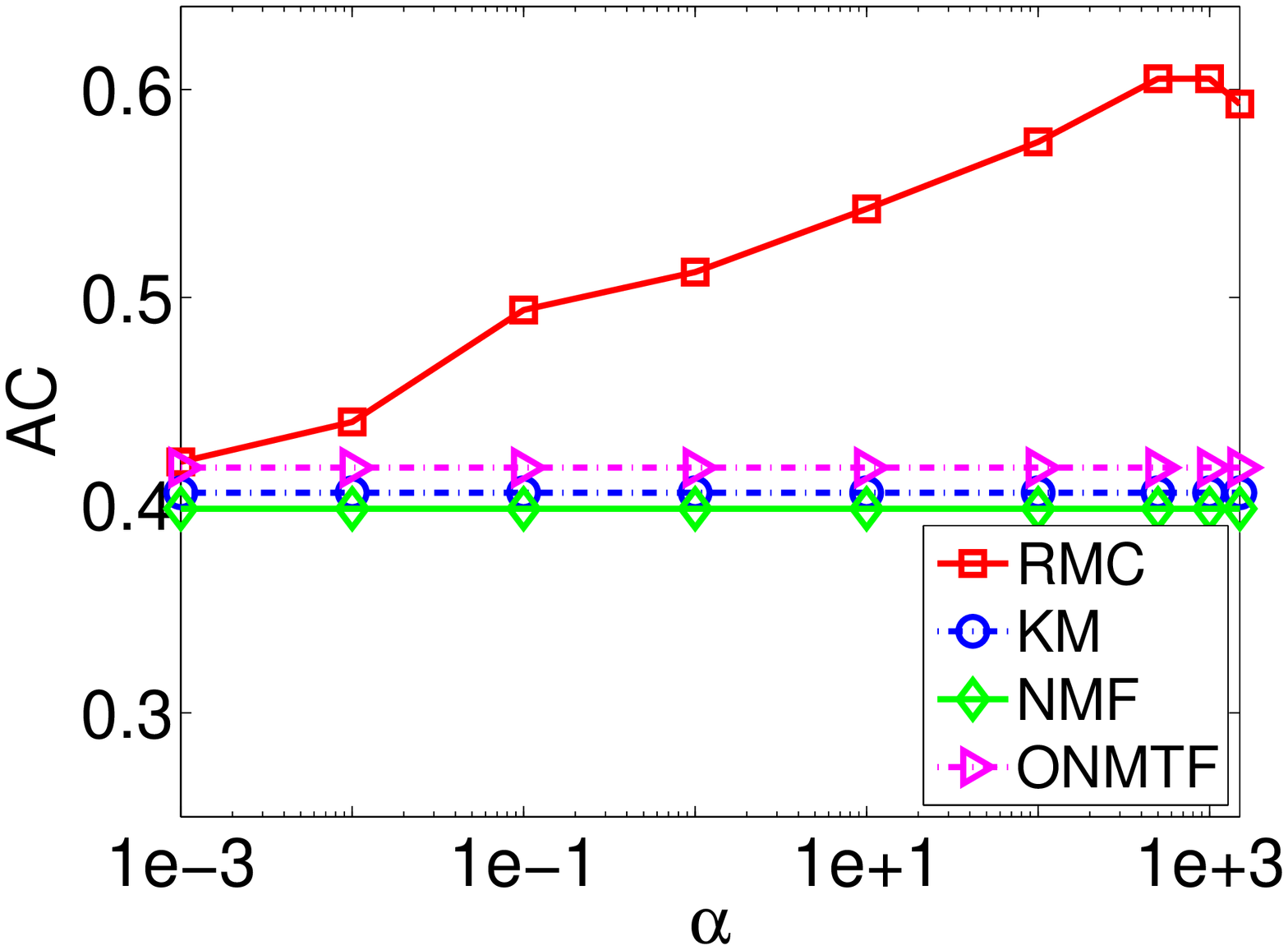}}
\subfigure[Leukemia2]{\includegraphics[width=0.23\textwidth]{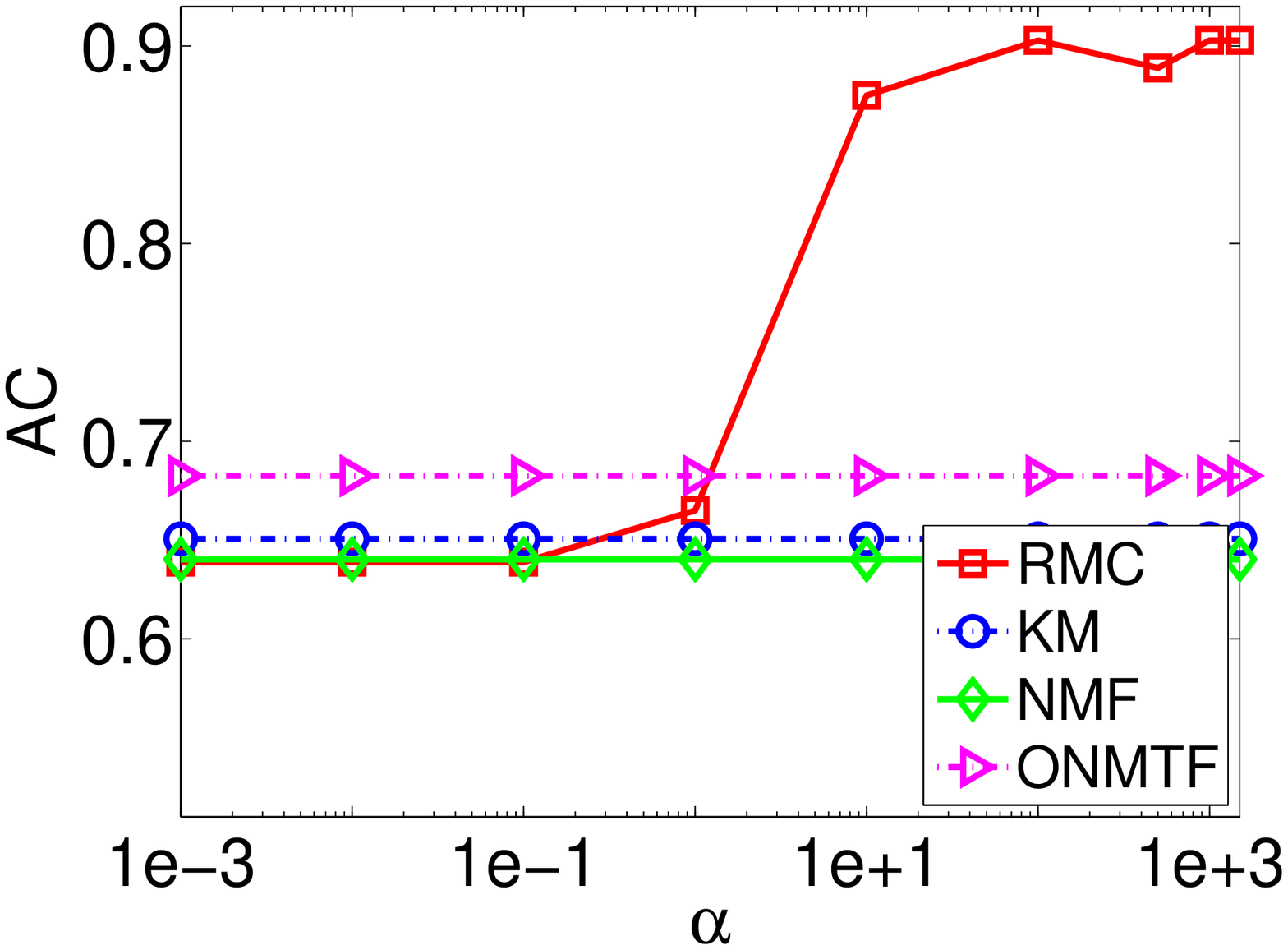}}
\subfigure[LungCancer]{\includegraphics[width=0.23\textwidth]{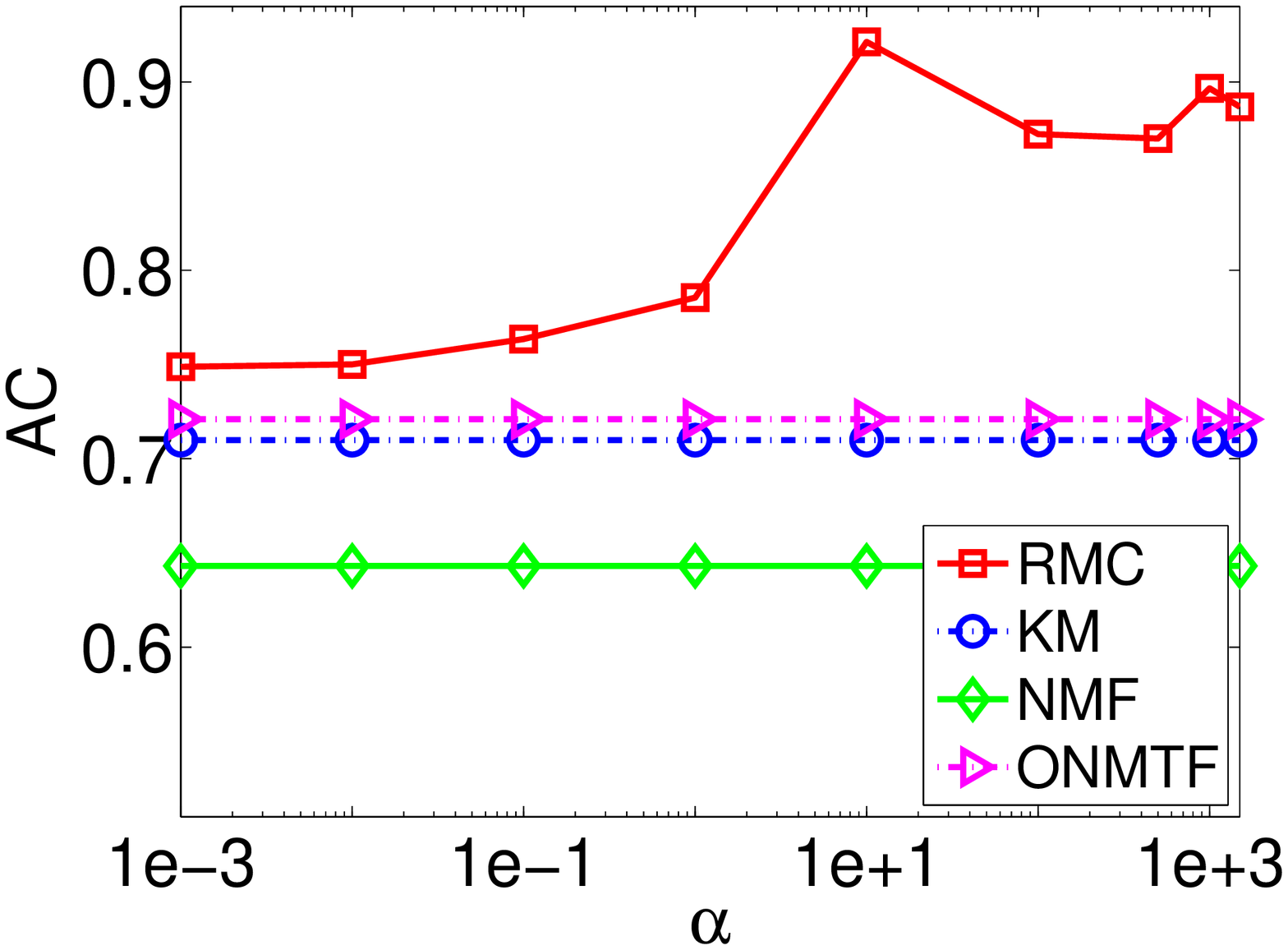}}
\subfigure[SRBCT]{\includegraphics[width=0.23\textwidth]{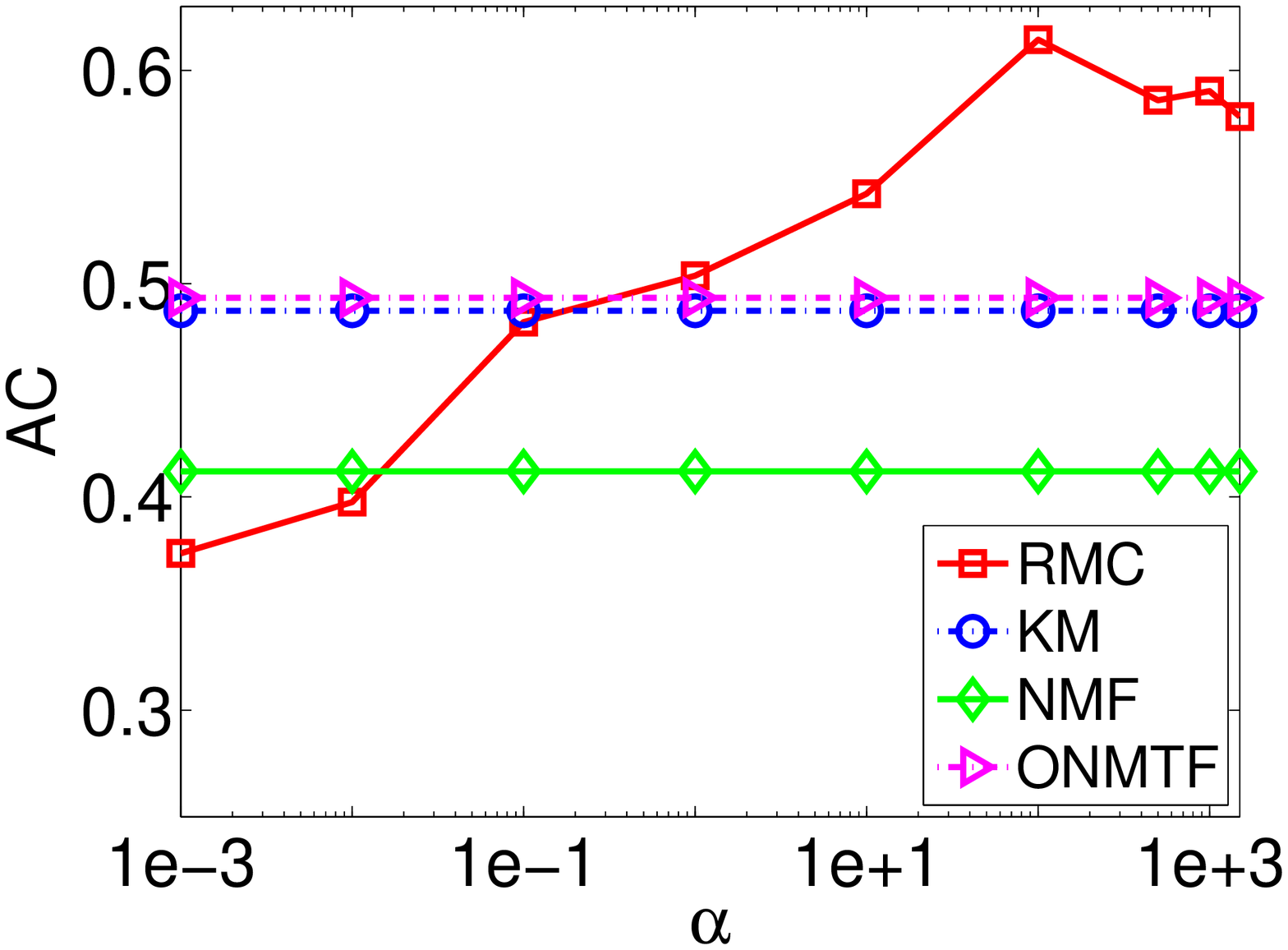}}
\caption{Parameter selection results in terms of clustering accuracy with varied $\alpha$ of RMC on different data sets.}
\label{fig:alphaSelAC} 
\end{figure*}
%
\begin{figure*}[!ht]
\centering
\subfigure[NGroups5]{\includegraphics[width=0.23\textwidth]{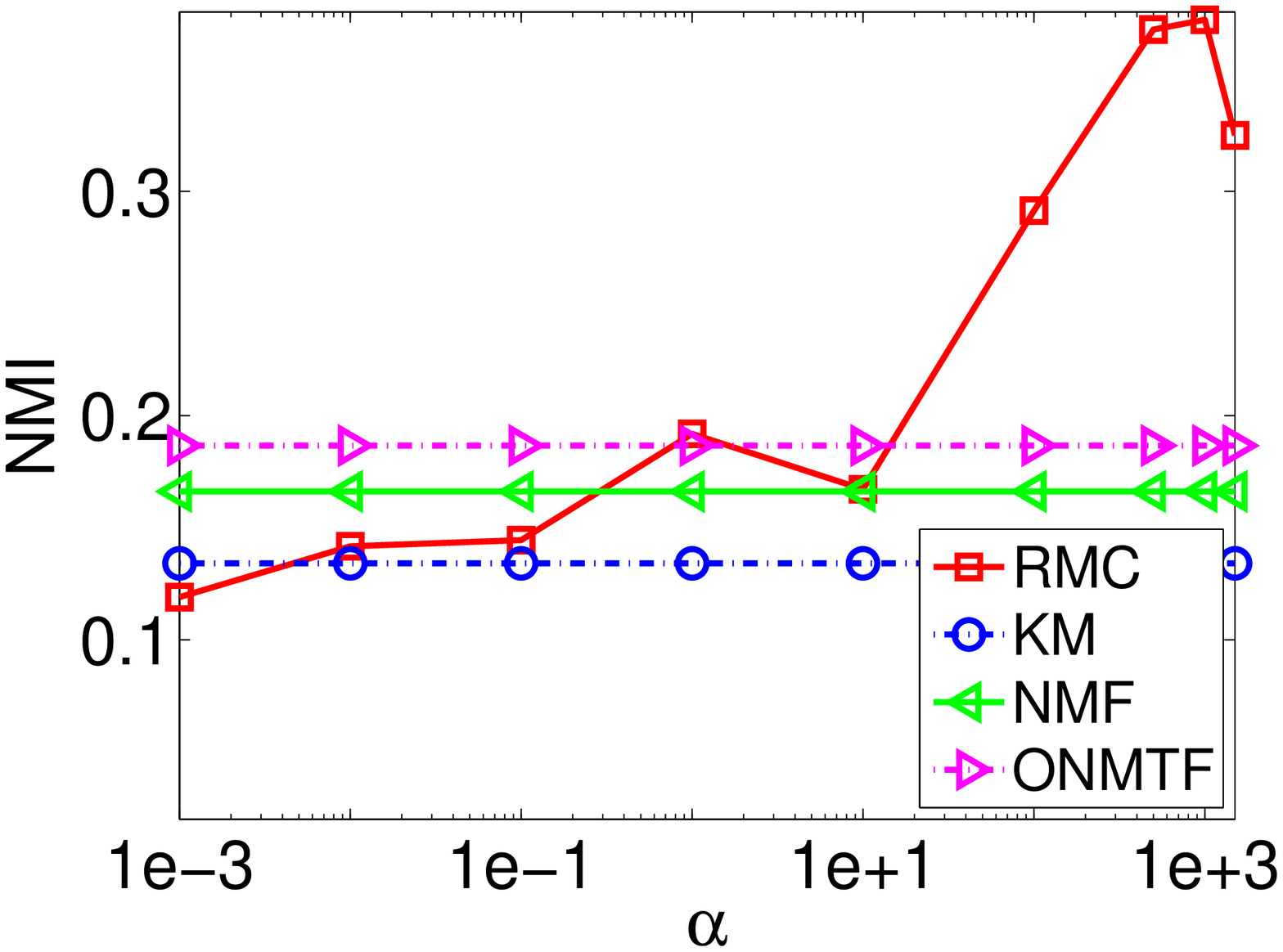}}
\subfigure[RCV1-5]{\includegraphics[width=0.23\textwidth]{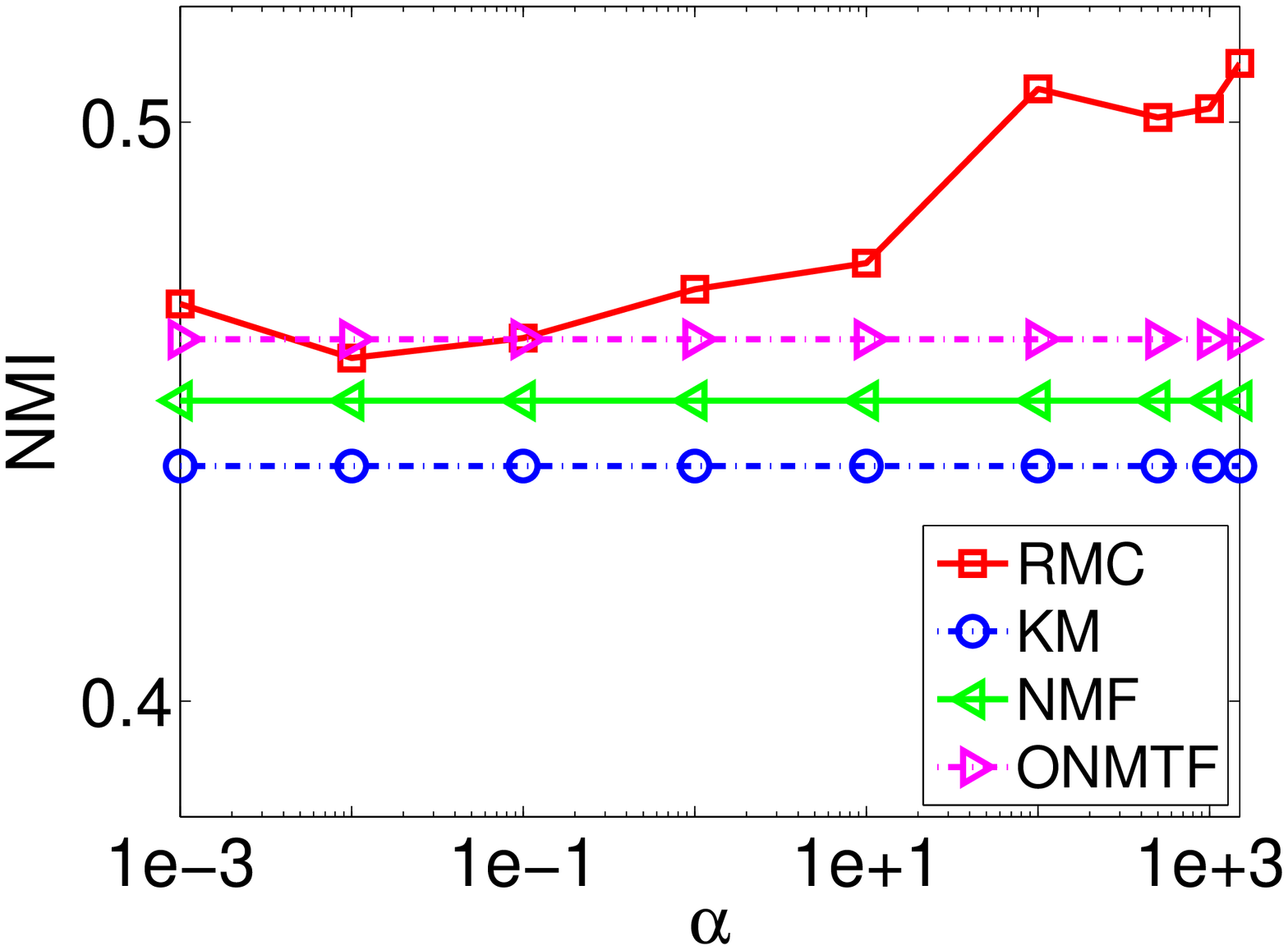}}
\subfigure[COIL20]{\includegraphics[width=0.23\textwidth]{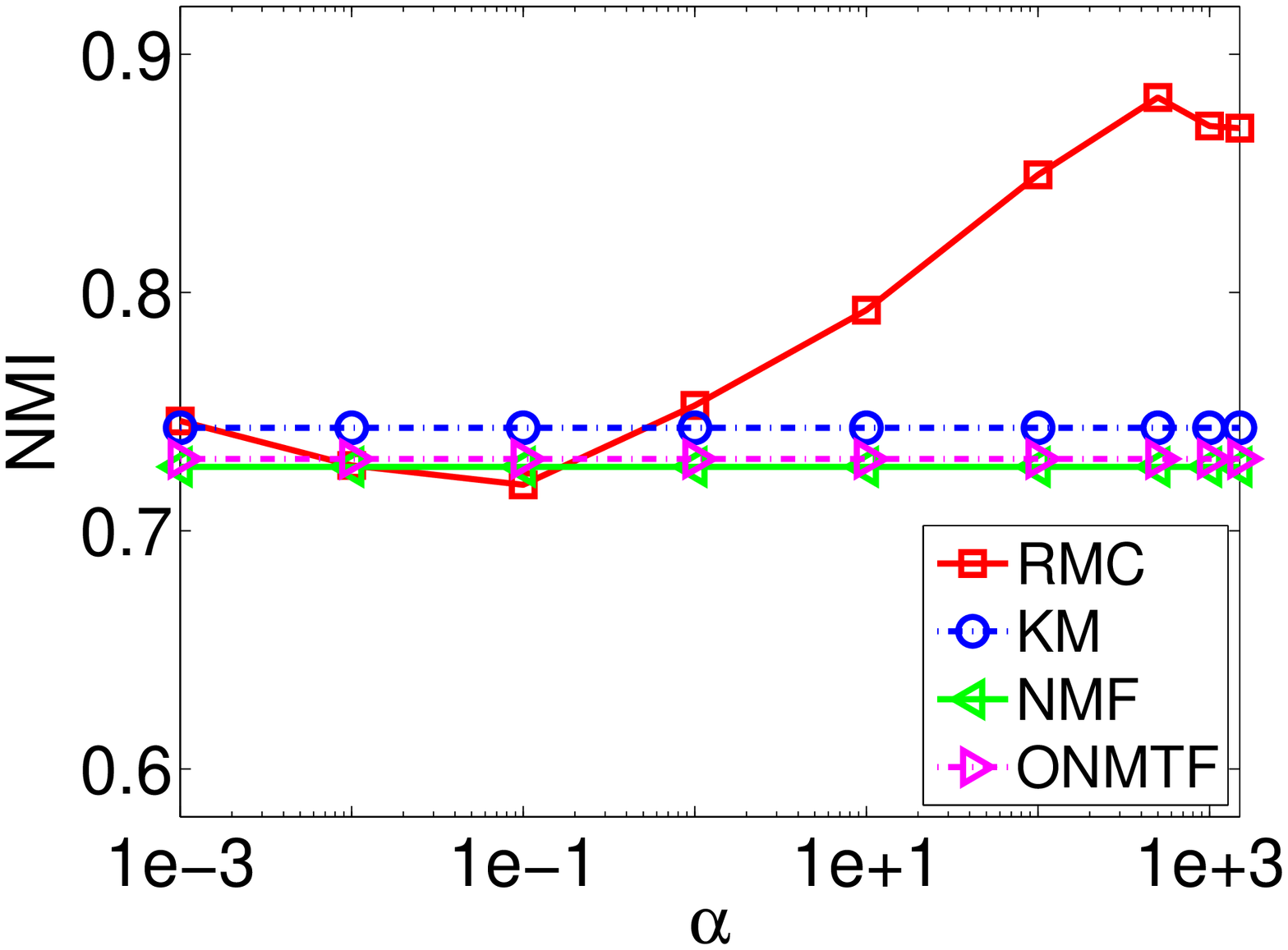}}
\subfigure[AlphaDigit]{\includegraphics[width=0.23\textwidth]{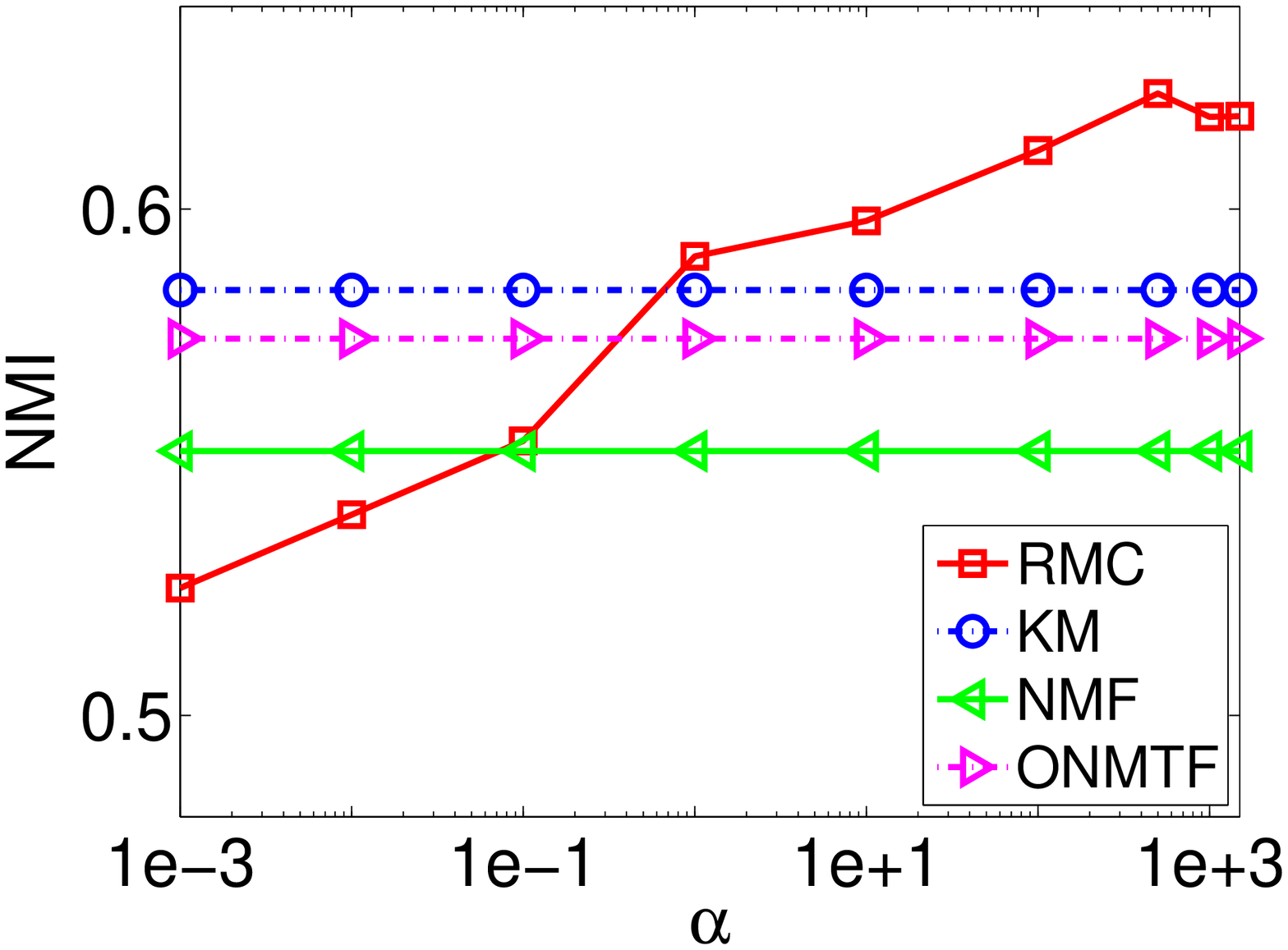}}
\subfigure[UMIST]{\includegraphics[width=0.23\textwidth]{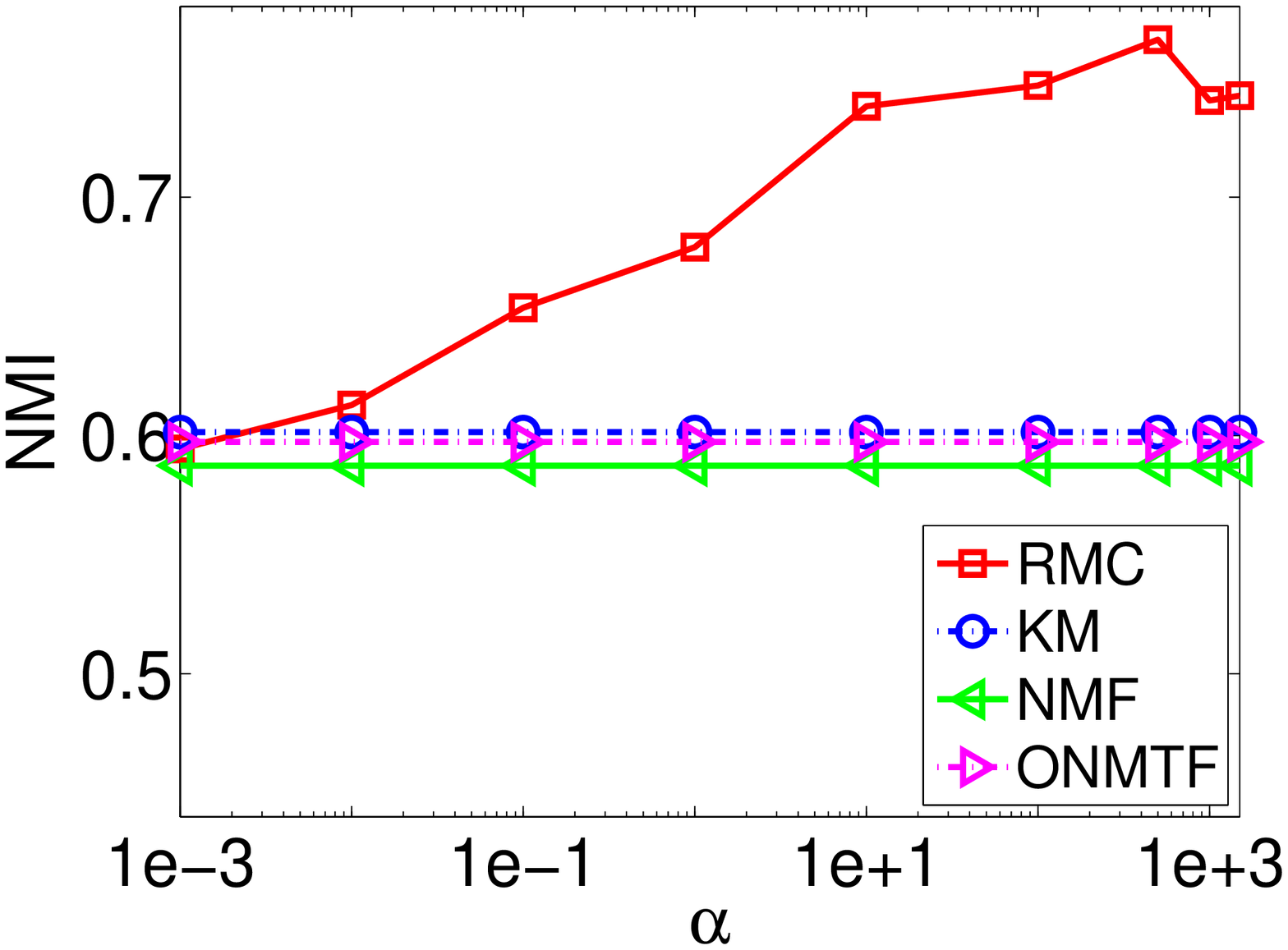}}
\subfigure[Leukemia2]{\includegraphics[width=0.23\textwidth]{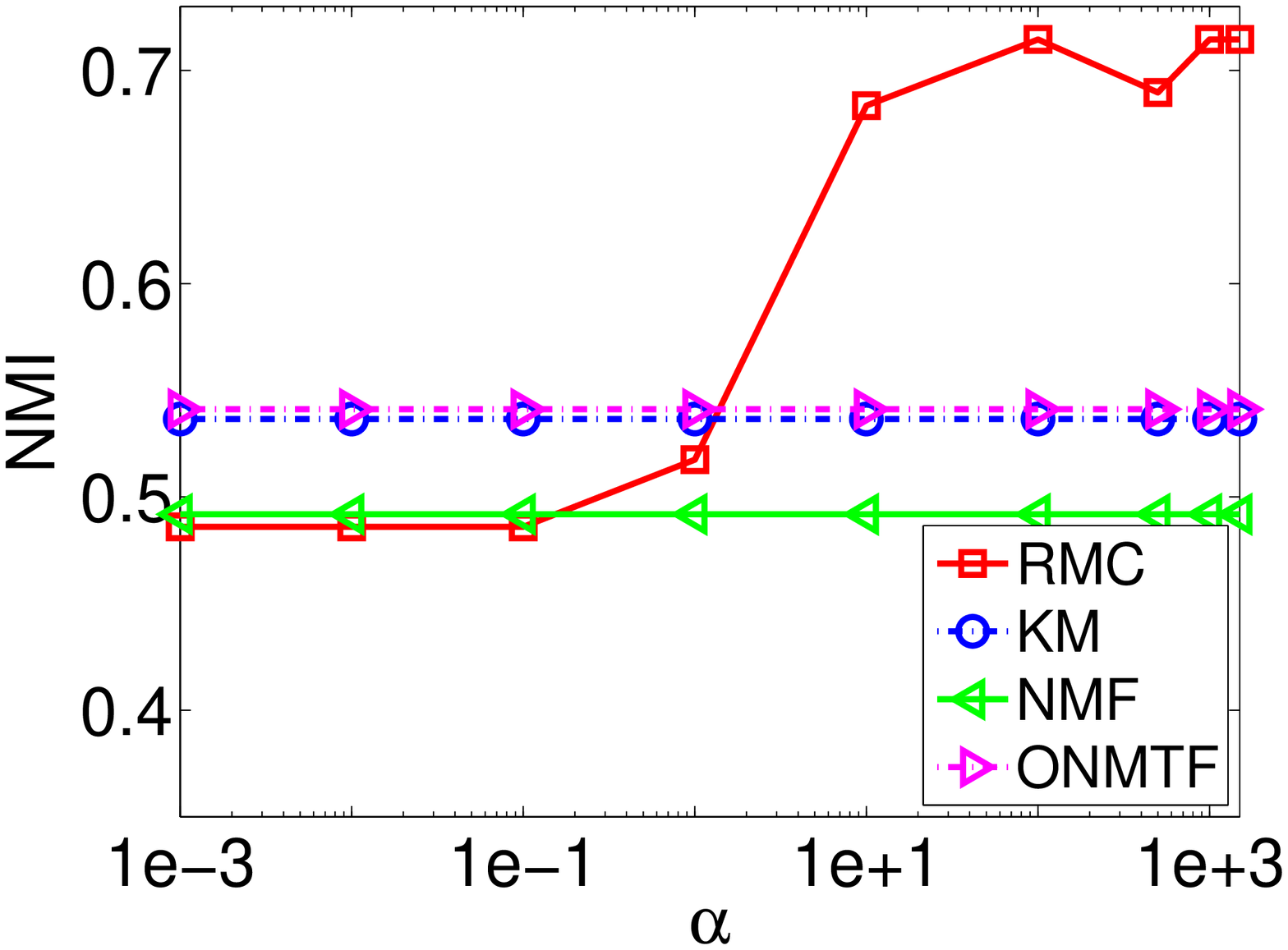}}
\subfigure[LungCancer]{\includegraphics[width=0.23\textwidth]{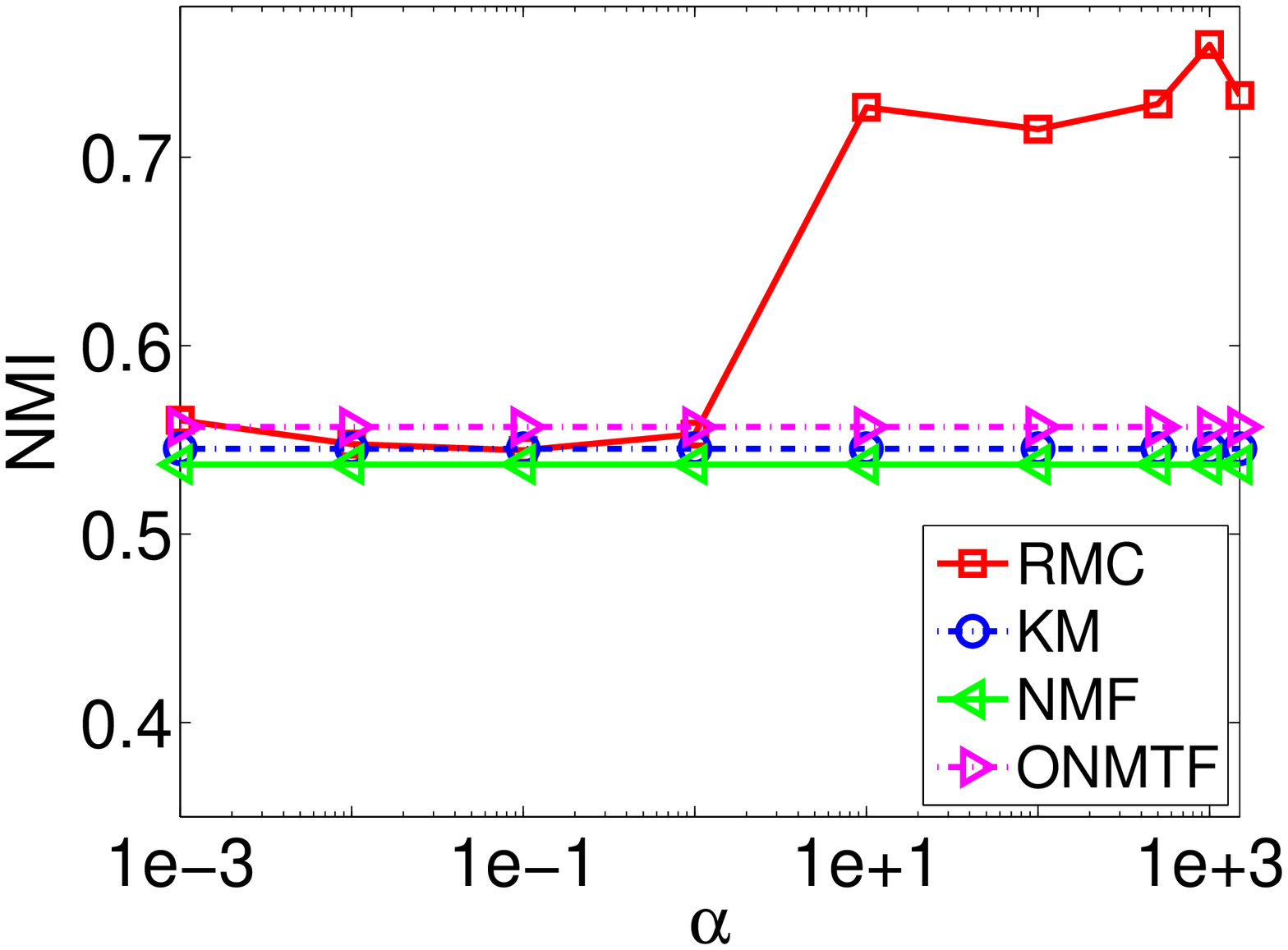}}
\subfigure[SRBCT]{\includegraphics[width=0.23\textwidth]{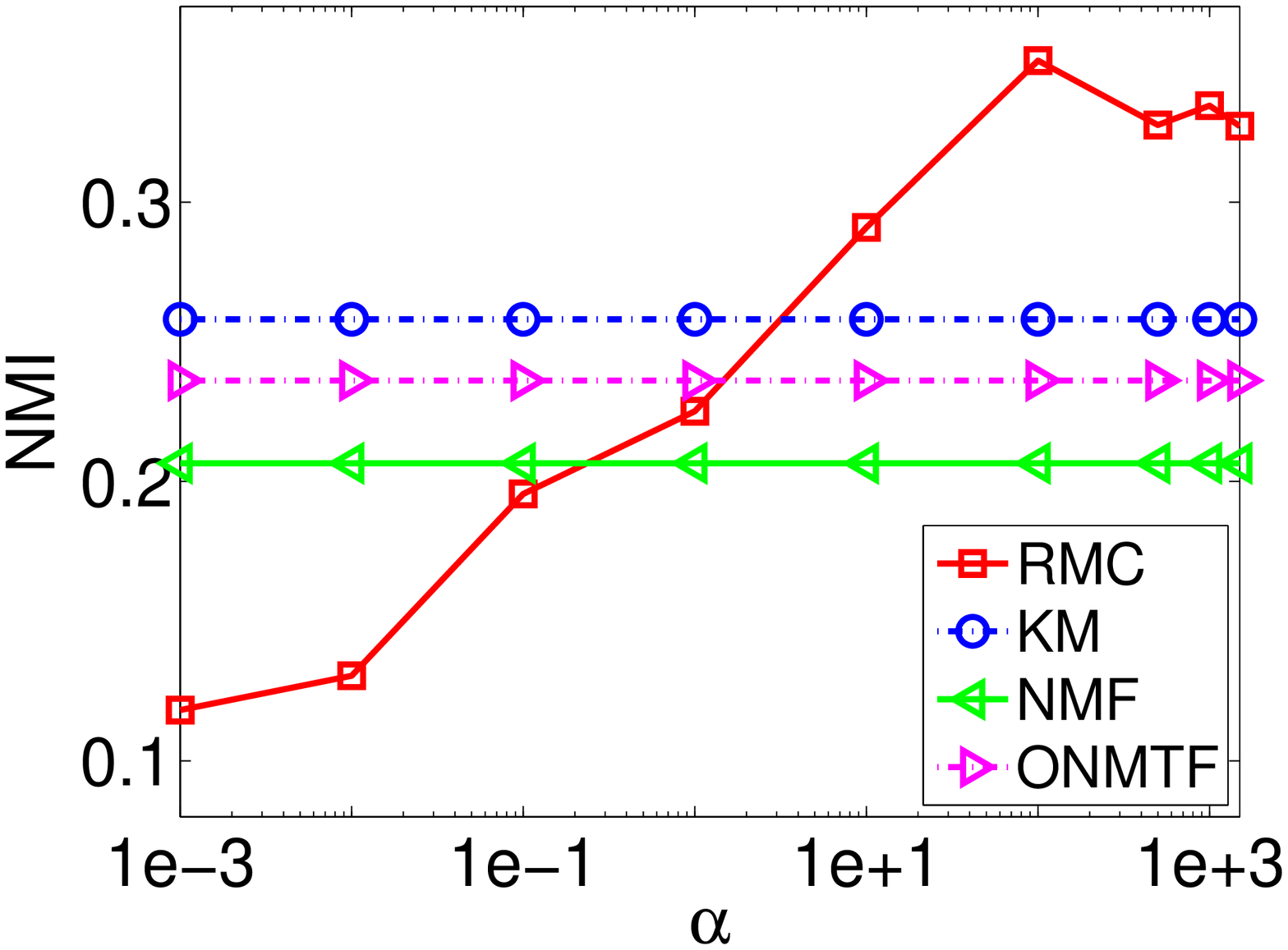}}
\caption{Parameter selection results in terms of normalized mutual information with varied $\alpha$ of RMC on different data sets.}
\label{fig:alphaSelNMI} 
\end{figure*}
%
\begin{figure*}[!ht]
\centering
\subfigure[NGroups5]{\includegraphics[width=0.23\textwidth]{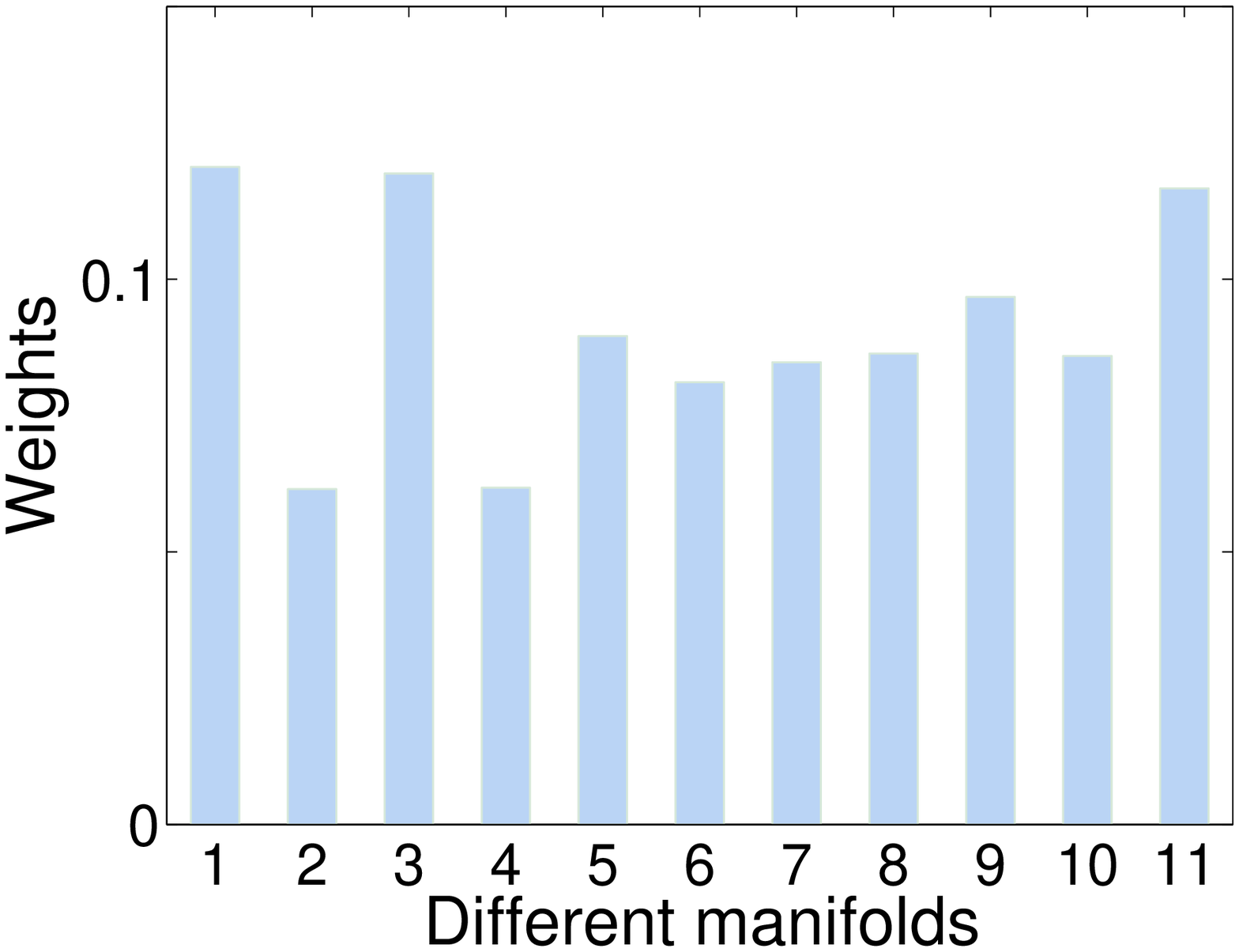}}
\subfigure[RCV1-5]{\includegraphics[width=0.23\textwidth]{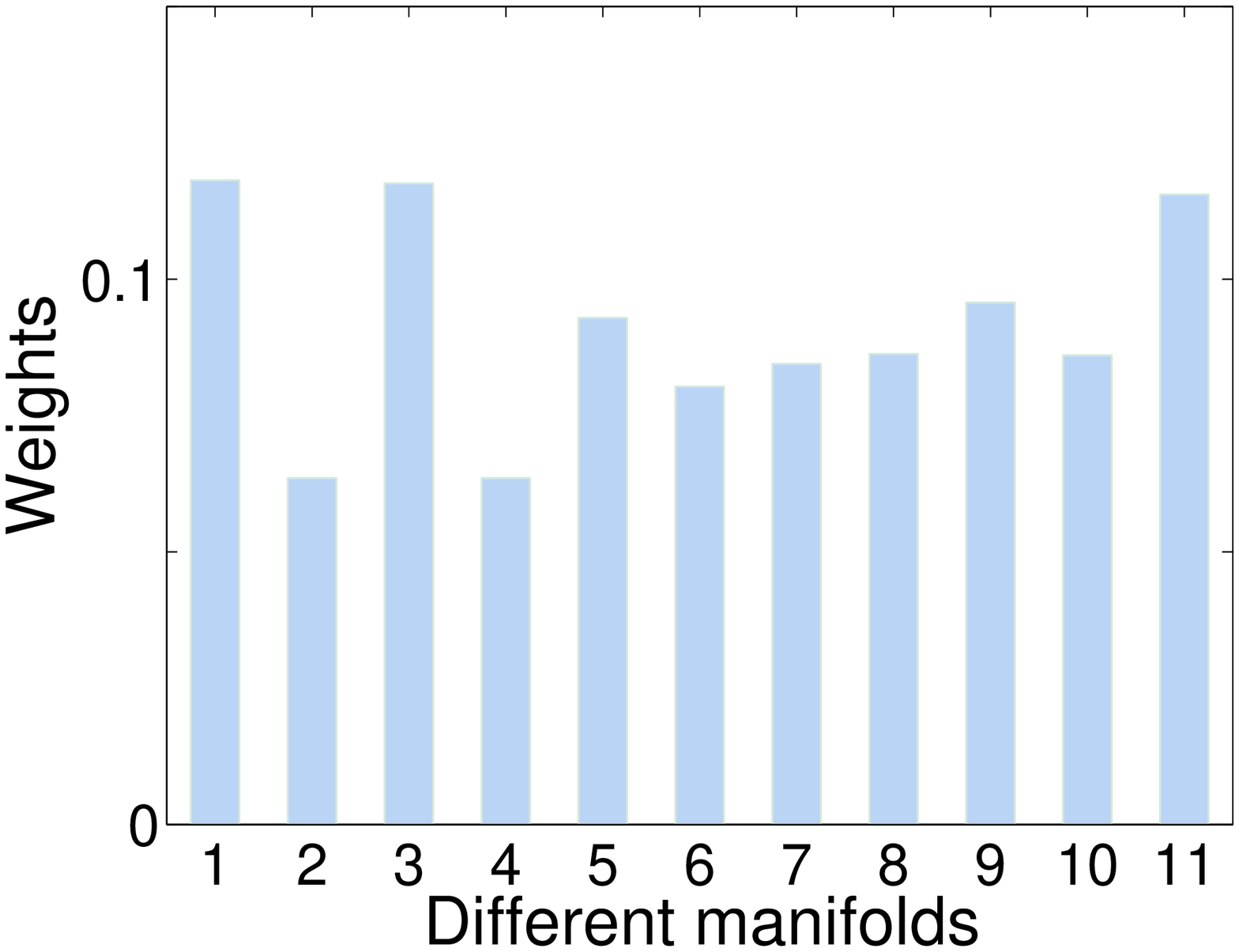}}
\subfigure[COIL20]{\includegraphics[width=0.23\textwidth]{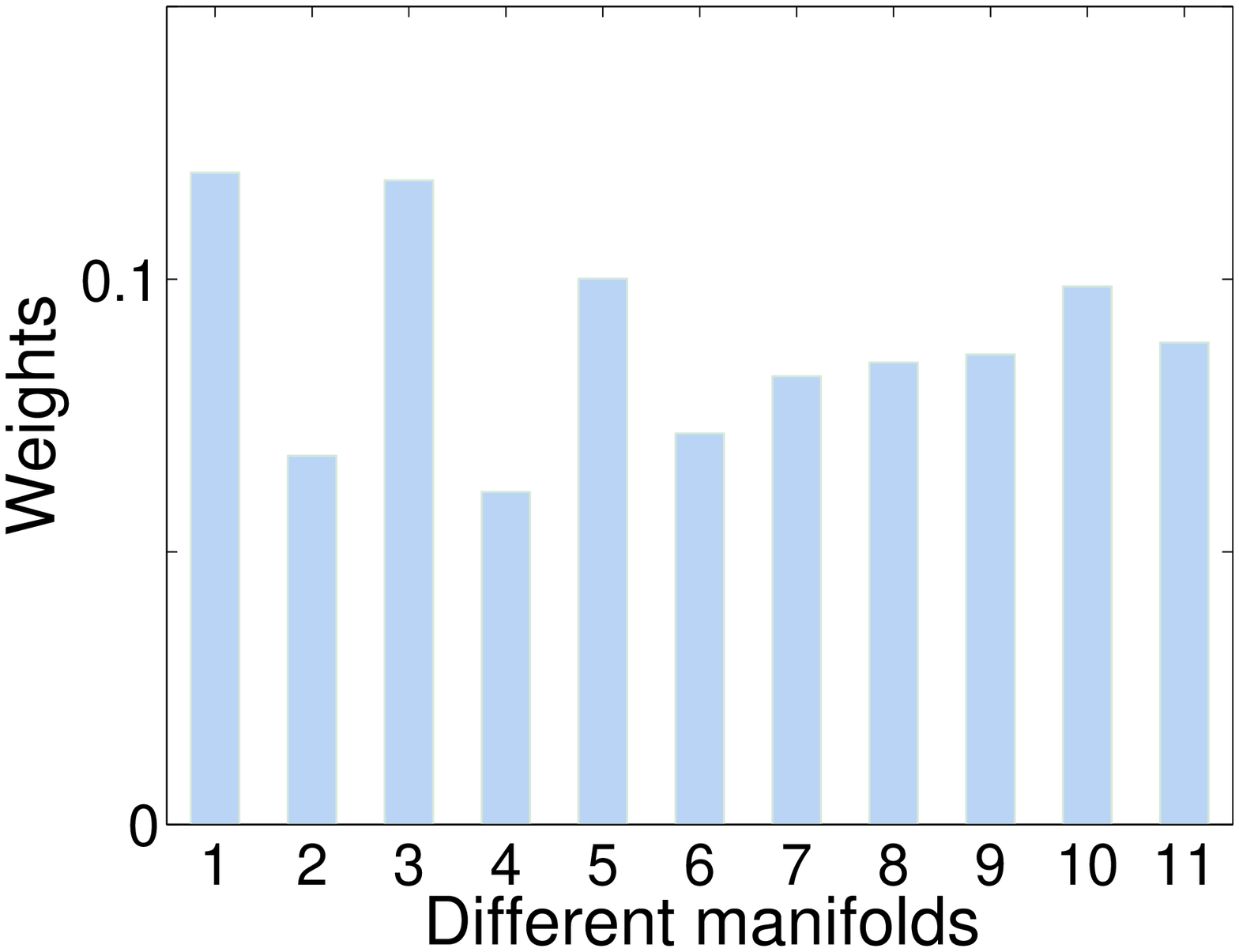}}
\subfigure[AlphaDigit]{\includegraphics[width=0.23\textwidth]{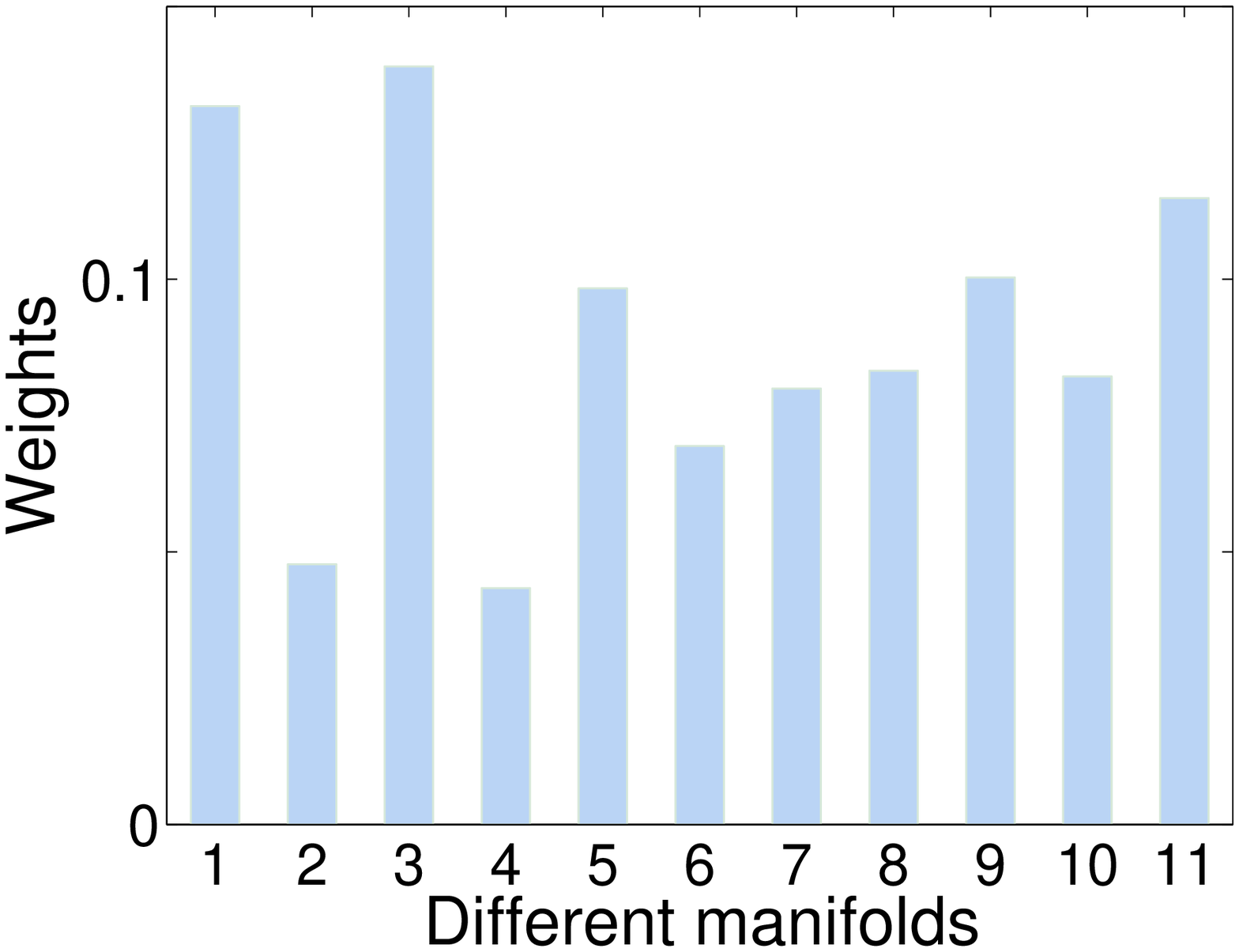}}
\subfigure[UMIST]{\includegraphics[width=0.23\textwidth]{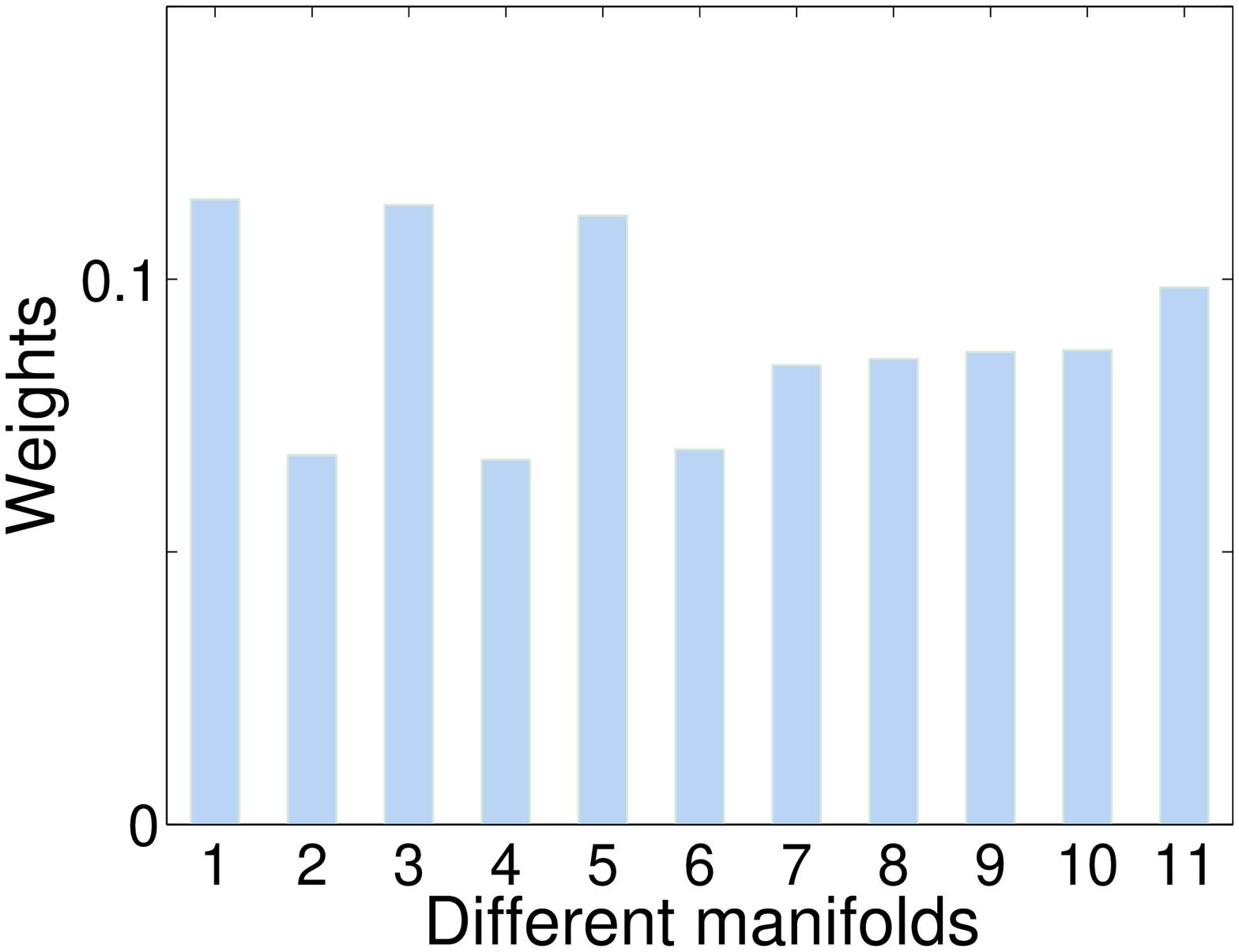}}
\subfigure[Leukemia2]{\includegraphics[width=0.23\textwidth]{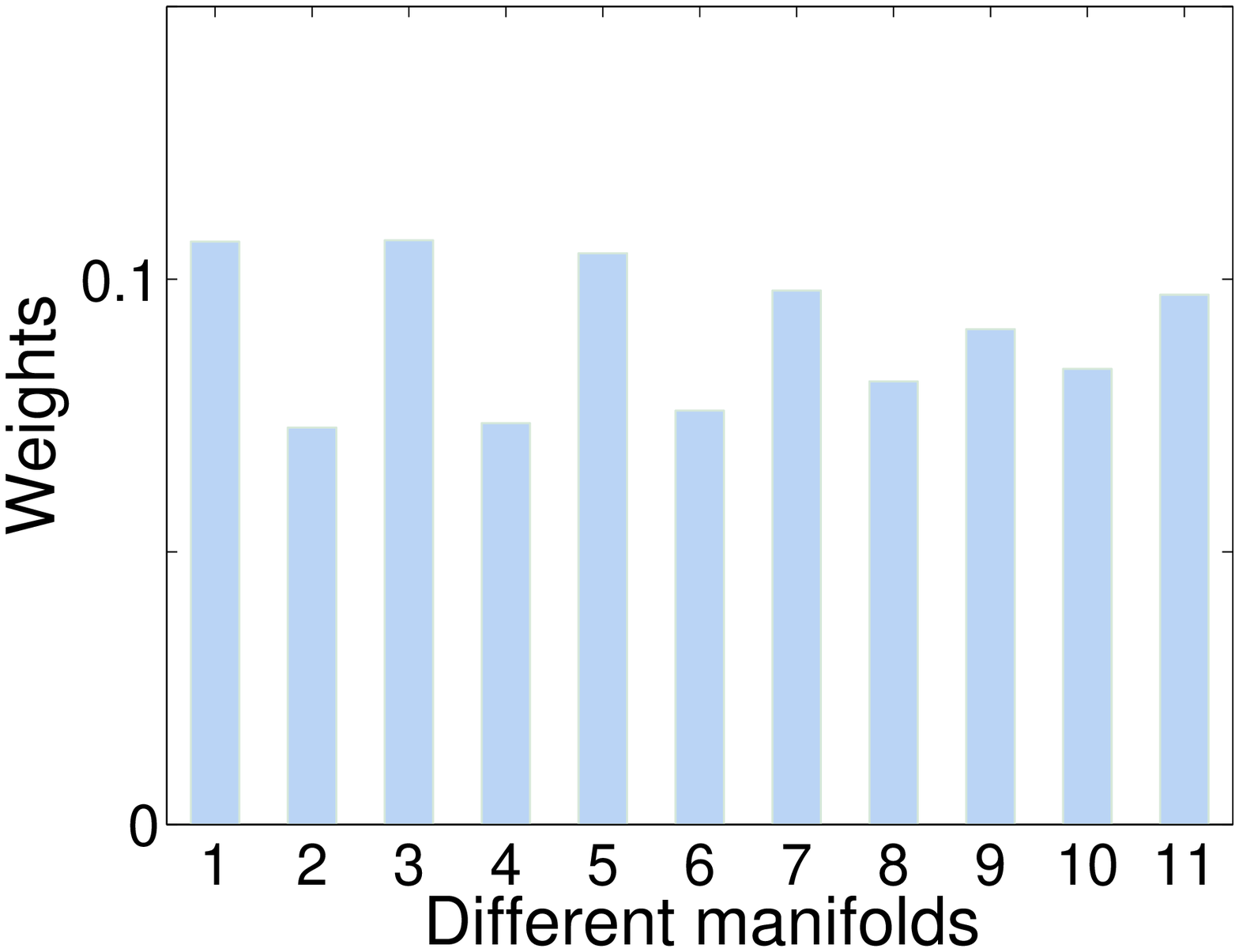}}
\subfigure[LungCancer]{\includegraphics[width=0.23\textwidth]{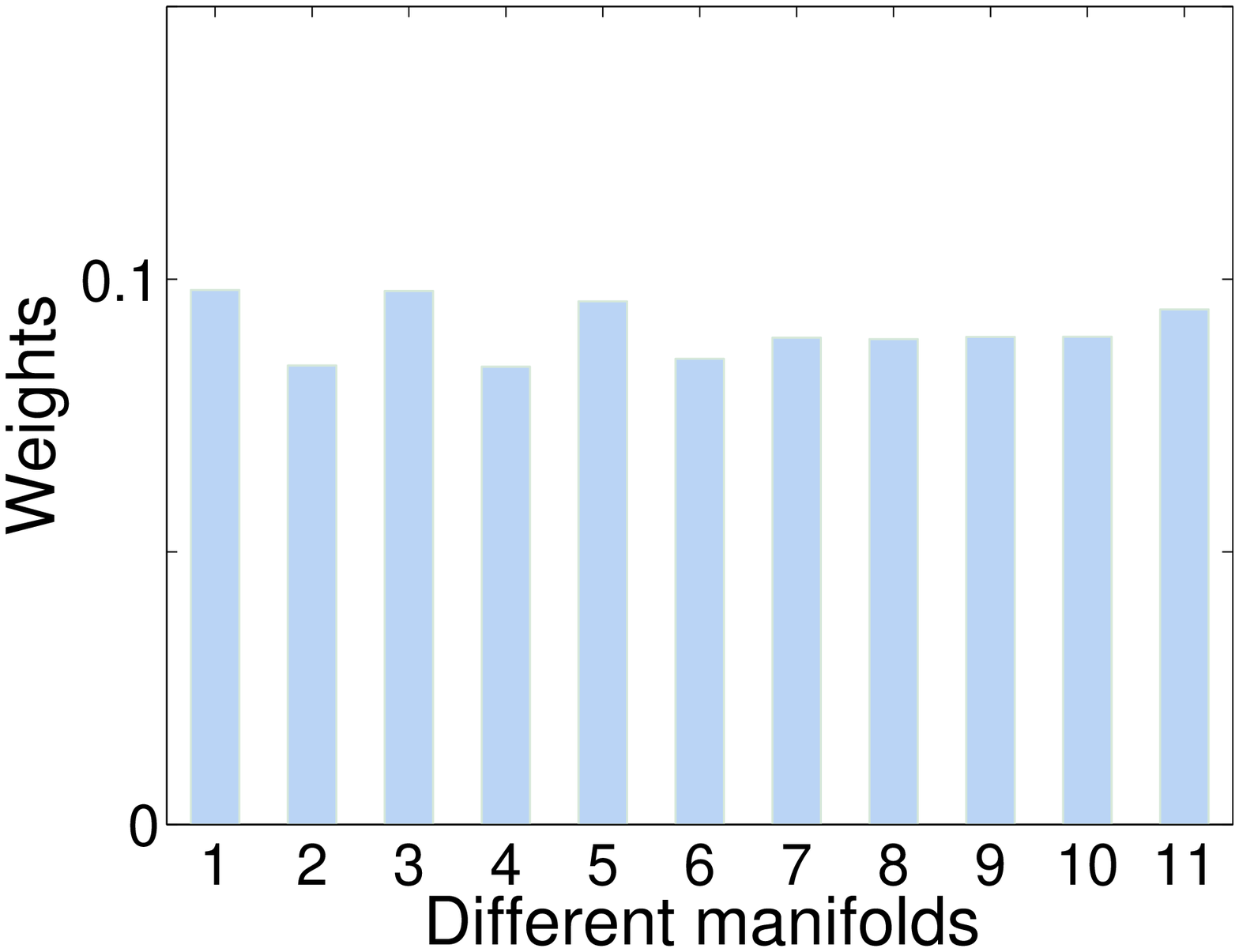}}
\subfigure[SRBCT]{\includegraphics[width=0.23\textwidth]{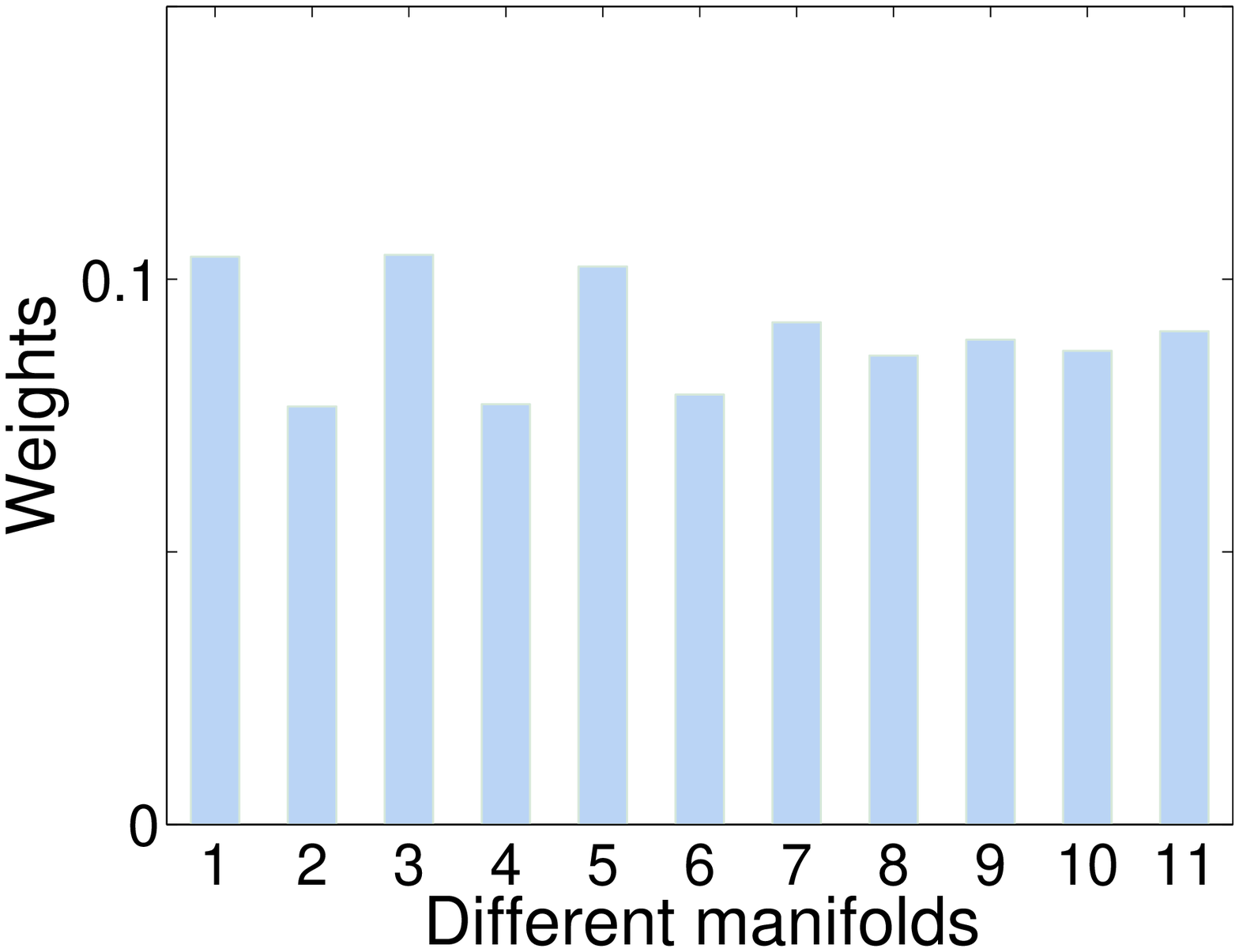}}
\caption{Histograms of the manifold coefficients obtained by using RMC-E on different data sets.}
\label{fig:muRMC-E} 
\centering
\subfigure[NGroups5]{\includegraphics[width=0.23\textwidth]{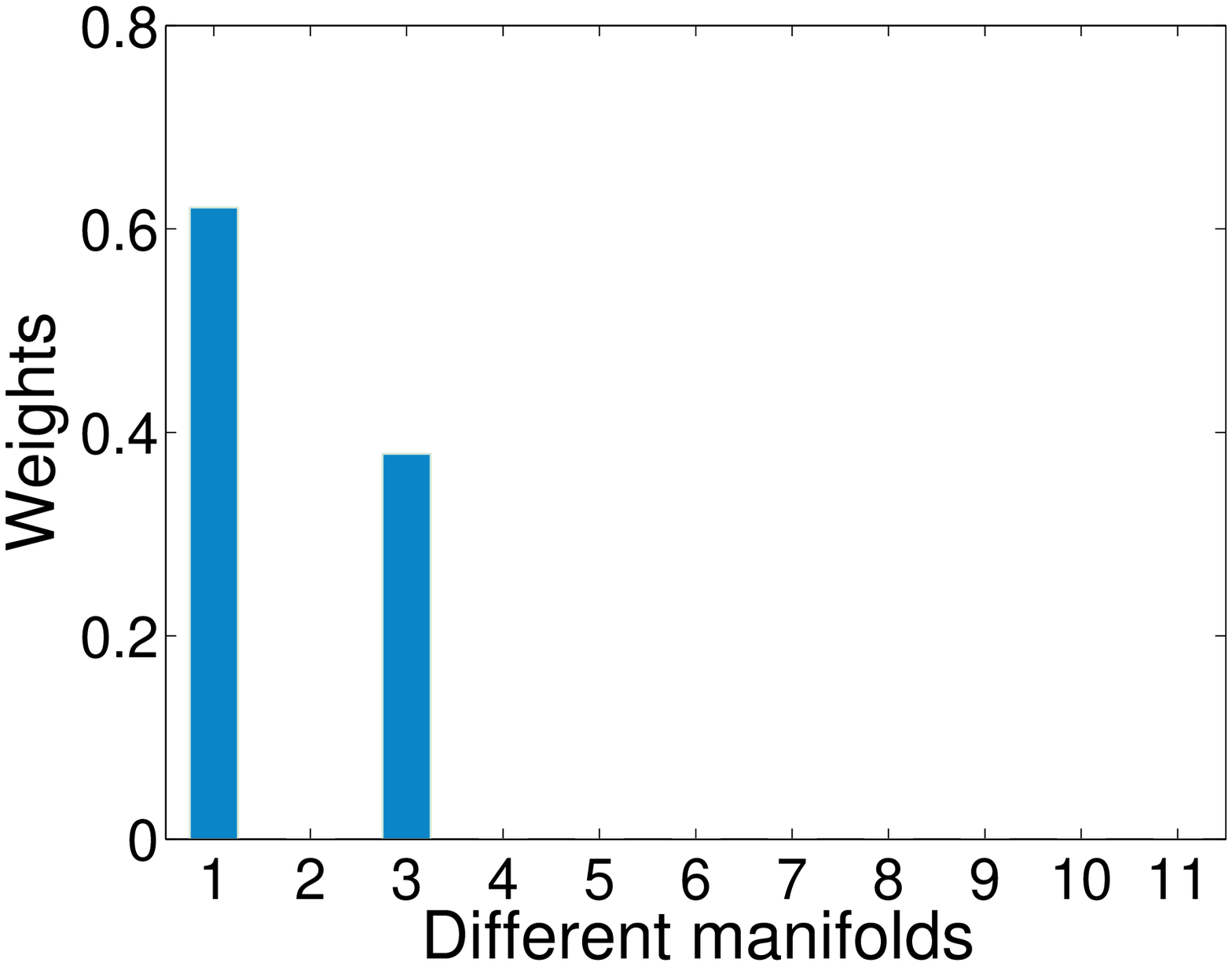}}
\subfigure[RCV1-5]{\includegraphics[width=0.23\textwidth]{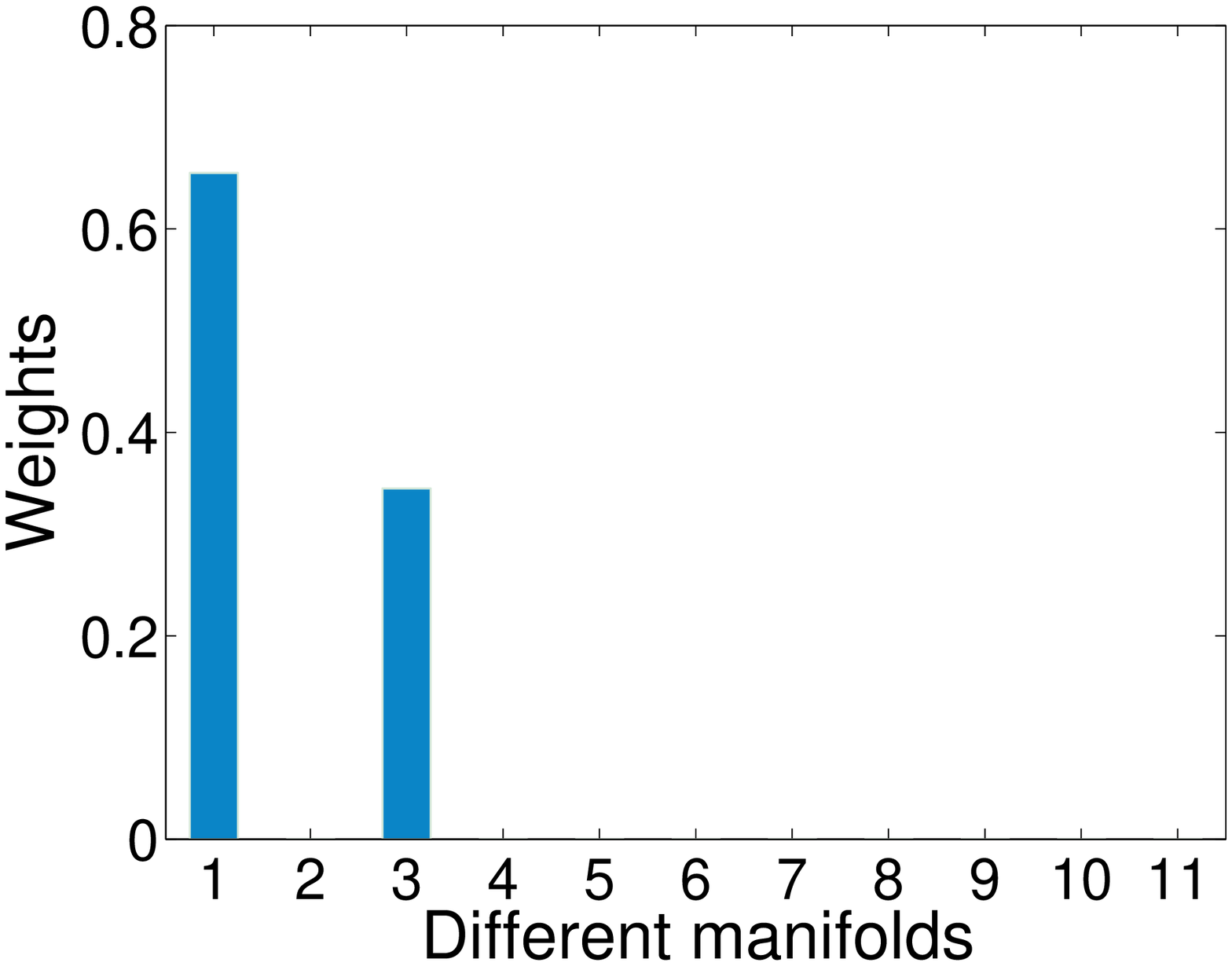}}
\subfigure[COIL20]{\includegraphics[width=0.23\textwidth]{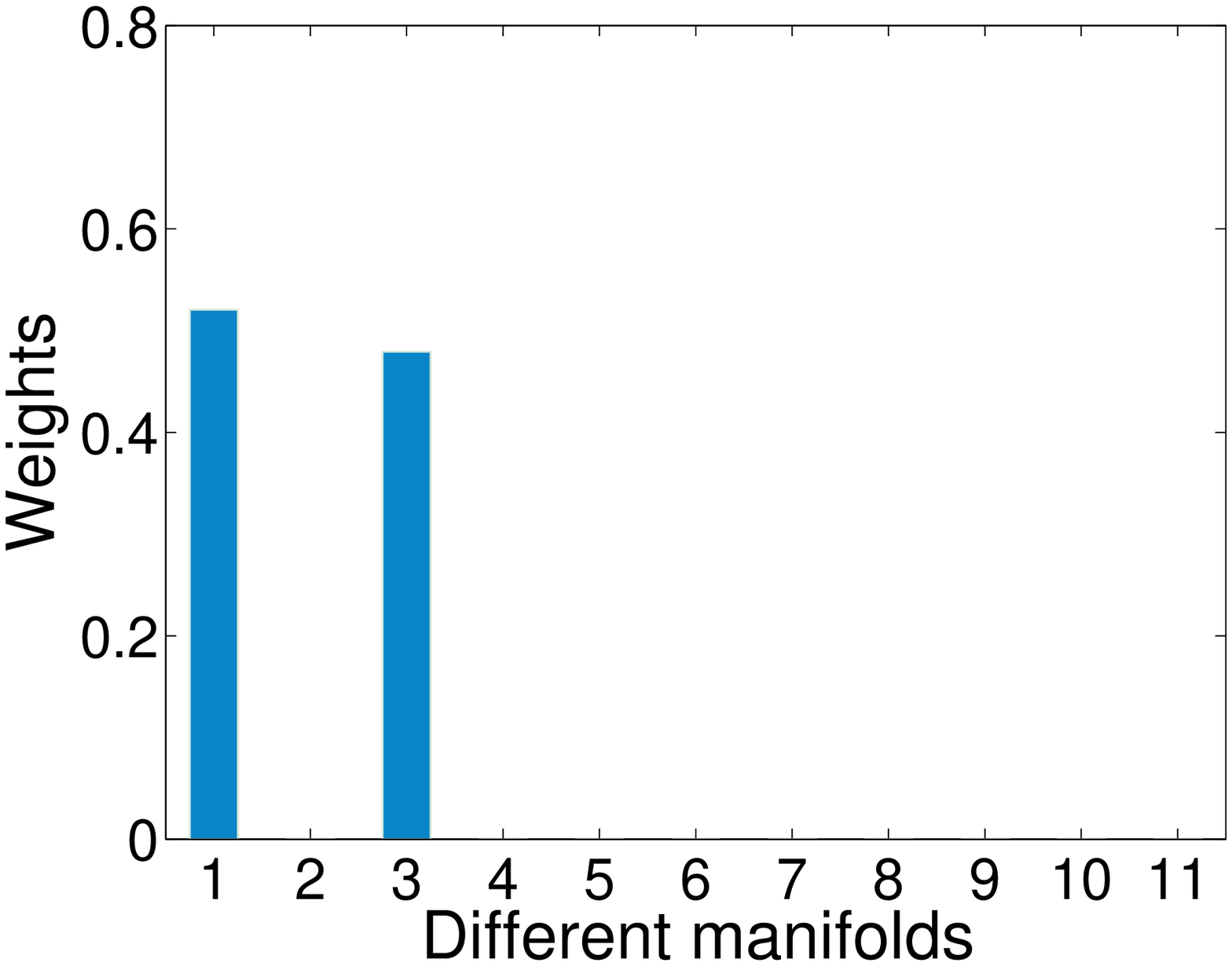}}
\subfigure[AlphaDigit]{\includegraphics[width=0.23\textwidth]{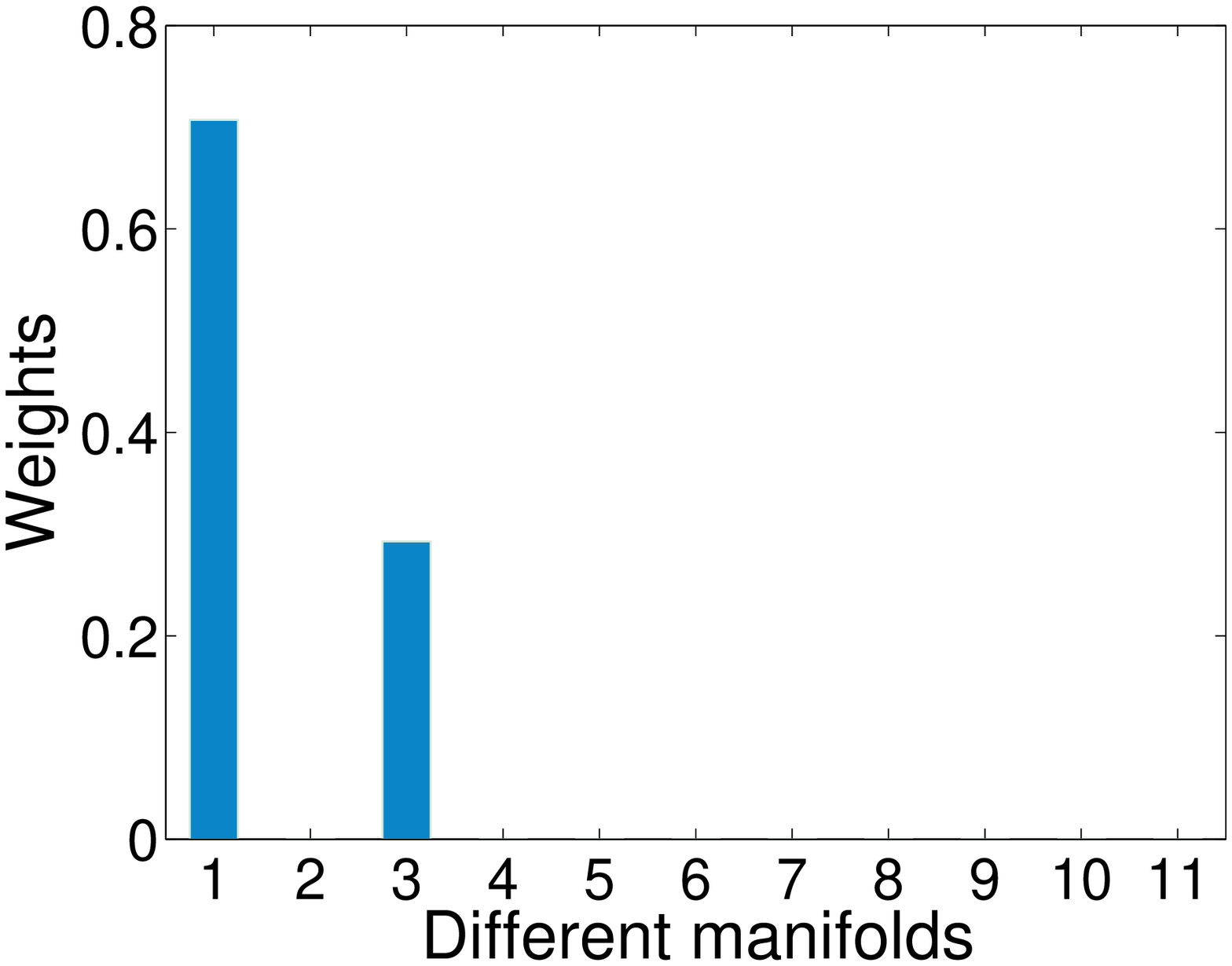}}
\subfigure[UMIST]{\includegraphics[width=0.23\textwidth]{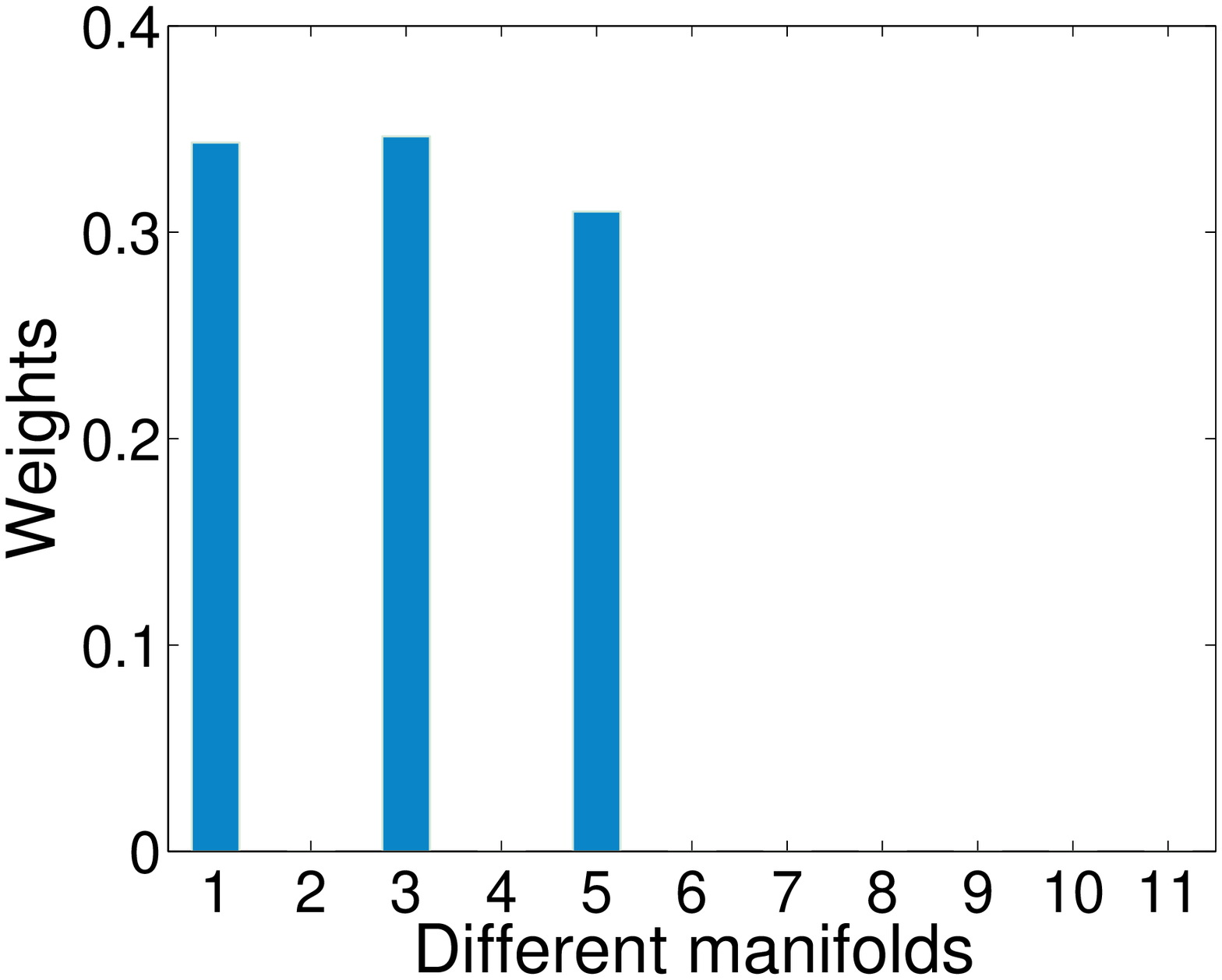}}
\subfigure[Leukemia2]{\includegraphics[width=0.23\textwidth]{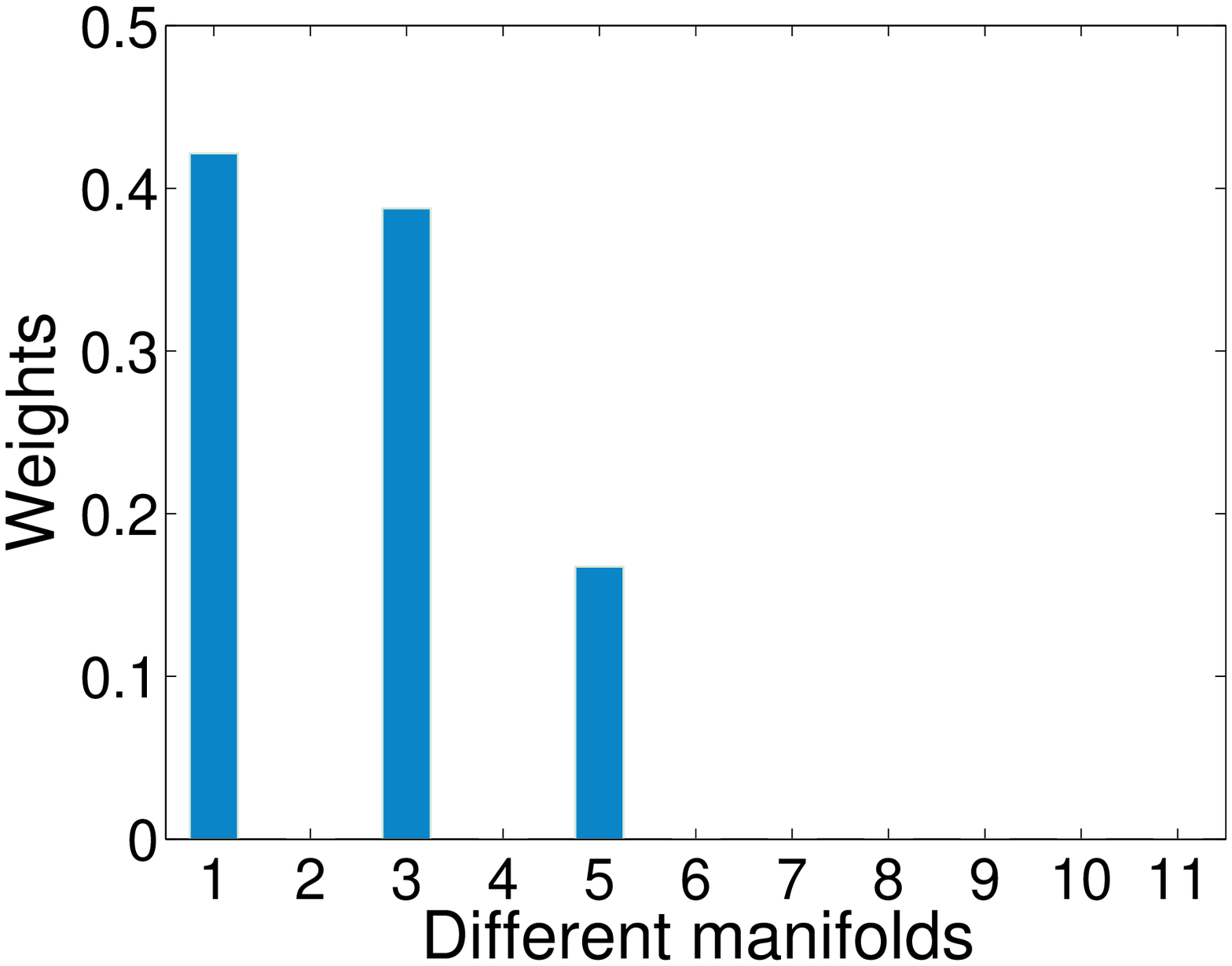}}
\subfigure[LungCancer]{\includegraphics[width=0.23\textwidth]{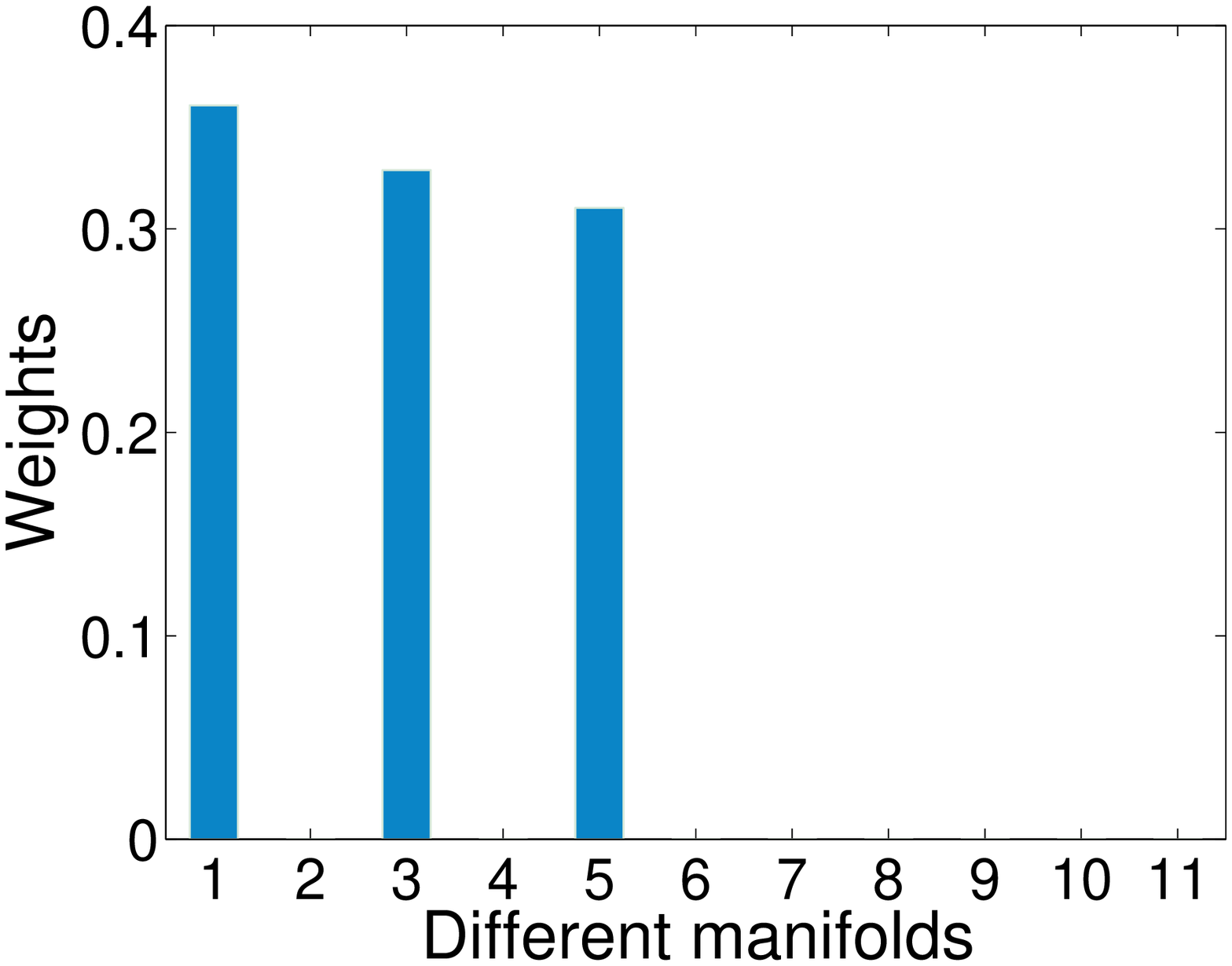}}
\subfigure[SRBCT]{\includegraphics[width=0.23\textwidth]{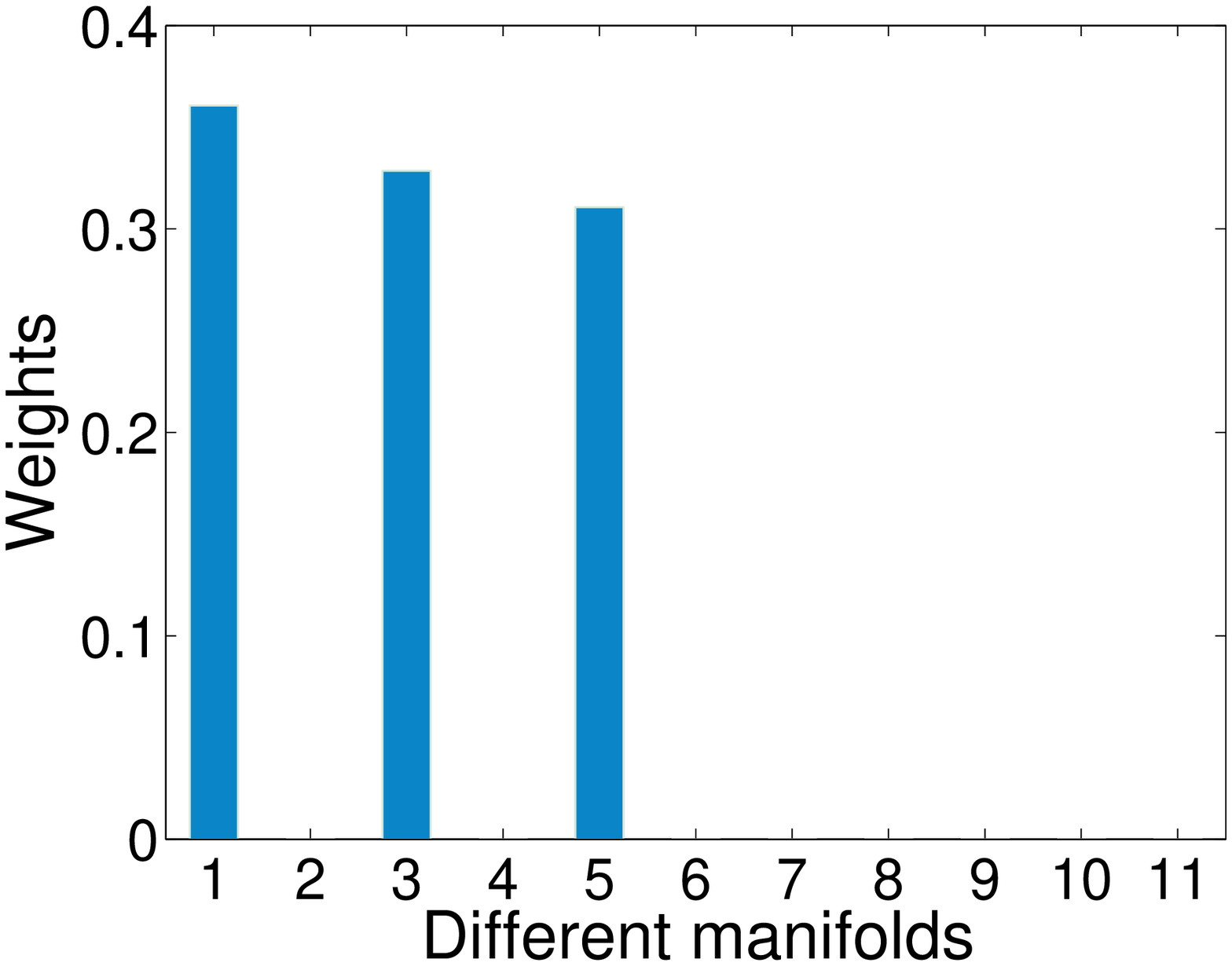}}
\caption{Histograms of the manifold coefficients obtained by using RMC-C on different data sets.}
\label{fig:muRMC-C} 
\end{figure*}

\subsection{Parameter Settings}
There are several parameters to be tuned in the evaluated algorithms. To ensure the fairness, we run them under different parameter settings, and report the best results.

For all methods except KM, the number of sample or feature clusters is set to the actual number of classes in all the collected data sets. Note that there is no parameter selection for KM, NMF and ONMTF, once the number of clusters is given. For the graph-based methods GNMF, DRCC, SNMTF, OSNTF and RMC, the number of the nearest neighbor is fixed to a small number 5 to ensure the locality preserving property. The regularization parameters are all searched from the grid $\{0.001,0.01,0.1,1,10,100,500,1500\}$. As for the co-clustering methods, the regularization parameters for the sample graph and the feature graph are set to the same. Except RMC, the other graph-based methods construct the graph Laplacian matrix using the binary weighting scheme as suggested in \cite{cai2011gnmf, gu2009coclu-graph}.

For our RMC approach, the ensemble dual regularization parameter $\alpha$ is in proportion to the over-fitting tolerance parameter $\beta$, which is empirically set to $\beta = 0.1\alpha$ \cite{geng2009emr}. To generate a group of diverse manifolds, we utilized three kinds of weighting schemes to construct the sample and feature graphs, \textit{i.e.}, the binary weighting, the Gaussian kernel, and the cosine similarity. In particular, for the Gaussian kernel, we varied the bandwidth $t$ in a broad range of area, \textit{i.e.}, $t = \{\frac{\tau}{100}, \frac{\tau}{60}, \frac{\tau}{30}, \frac{\tau}{10}, \tau, 10\tau, 30\tau,  60\tau, 100\tau \}$. Here, $\tau$ is empirically set as the inverse of the mean square of Euclidean distances between all sample or feature pairs in the selected data \cite{geng2009emr}, \textit{i.e.}, $\tau = \bigl(\frac{1}{n^2}\sum_{i,j=1}^n \|\bm{x}_i- \bm{x}_j\|^2\bigr)^{-1}$. In total, \textit{eleven} manifolds containing \textit{nine} Gaussian graphs, \textit{one} binary graph and \textit{one} cosine similarity graph were employed in manifold ensemble learning for the sample and feature spaces.
\subsection{Results}
The averaged results of different algorithms are tabulated in Table~\ref{table:AC} and \ref{table:NMI} in terms of AC and NMI, respectively. The best results on each data set are highlighted in boldface. Besides, the asterisk symbol "\textbf{*}" aside our results indicates RMC is statistically and significantly better than the other well-established methods at a significance level of $0.01$.

From these experiments, several interesting points can be revealed below.
\begin{itemize}
\item The clustering performances of RMC are systematically and consistently better than the compared algorithms, which verifies that the manifold ensemble learning is advantageous to the graph-based symmetric nonnegative matrix factorization methods for co-clustering.

\item Except the gene expression data, RMC-E generally outperforms RMC-C on the remaining data sets. We attribute this to the fact that they respectively employ two different optimization methods to learn the manifold coefficients. Specifically, EMDA has a global efficiency estimate mildly dependent on the number of manifolds for the convex optimization over the unit simplex, and is provably very efficient for large scale problems \cite{beck2003mirror}. On the contrary, in each iteration of CDA, only two variables are selected to update while the others are fixed, which is a pairwise alternating optimization, leading to sub-optimal solutions.

\item Generally, both RMC-E and RMC-C perform more or less on the gene expression data, which might be due to the reason that their sample sizes are much smaller than those of text and image data sets, \textit{i.e.}, \textit{small sample size} (SSS) problem. Thus, both EMDA and CDA are very likely to converge at a value around the optimal solution of the convex optimization problem, thereby achieving similar co-clustering results.

\item The graph regularization based methods, \textit{e.g.}, GNMF, DRCC, SNMTF, OSNTF and RMC, almost perform better than KM, NMF and OSNMTF. This is mainly for the reason that the graph-based approaches have considered the local geometrical structure, which is conducive to preserve the locality property in the low-dimensional data space.

\item The dual regularization based methods, \textit{e.g.}, DRCC and RMC, generally outperform the one-side graph regularization based method GNMF. This demonstrates that the geometric structures in both the sample and feature spaces are beneficial to further improve the clustering performance.

\end{itemize}

\subsection{Parameter Selection}
There are three important parameters in our RMC method, \textit{i.e.}, the manifold regularization parameter $\alpha$, the over-fitting tolerance parameter $\beta$ and the number of nearest neighbor $p$. Since the number of nearest neighbor has been fixed to 5 for all the graph-based approaches and no bias will occur during the comparison, it makes sense that here we neglect the model selection on the parameter $p$. As mentioned earlier in parameter settings, the over-fitting tolerance parameter $\beta$ was empirically set as $\beta = 0.1 \alpha$ \cite{geng2009emr} to reduce the degree of the parameter freedom in the objective function of RMC. Hence, it is only desirable to explore the influence of the parameter $\alpha$ on the clustering performance of our method. To do this, we varied the parameter $\alpha$ in a broad real value range, \textit{i.e.}, $\{0.001,0.01,0.1,1,10,100,500,1500\}$.

Figure~\ref{fig:alphaSelAC} and \ref{fig:alphaSelNMI} respectively show the plots of accuracy and normalized mutual information versus the parameter $\alpha$ for RMC using EMDA. Similar tendencies can be also observed for RMC using CDA. As vividly depicted in these figures, we can see that our approach enjoys the satisfactory performance when the parameter $\alpha$ takes a higher value, \textit{e.g.}, around 500 or 1000. This indicates that the ensemble manifold regularization term should be imposed larger weights, such that it can make much more positive contributions to the objective function of the proposed approach.
\subsection{Study on Manifold Coefficients}
Since manifold ensemble learning plays an essential role in the proposed RMC method, it is worthwhile to examine the manifold coefficients learned by two different optimization algorithms, \textit{i.e.}, EMDA and CDA. We use the histogram to draw the distribution of manifold coefficients derived from RMC under the best parameter settings. Figure~\ref{fig:muRMC-E} and \ref{fig:muRMC-C} respectively illustrate the histogram of manifold coefficients for RMC-E and RMC-C on all data sets.

From these bars, it is easy to find that CDA only choose two or three manifolds on most data sets, which suggests that there might be some important information loss, resulting that the improvement of the clustering performance seems not inspiring as shown in Table~\ref{table:AC} and \ref{table:NMI}. In contrast, EMDA assigns each manifold a nonzero coefficient or weight, and the convex combination of all the candidate manifolds jointly contributes to approximate the true intrinsic manifolds of the sample and feature spaces as much as possible, thus achieving greater performance improvements. Comparing the corresponding figures in RMC-E and RMC-C, we discover that the manifolds selected by RMC-C are the manifolds with the top two or three weights in RMC-E, which indicates that the two optimization methods almost select the similar manifolds as their main components. However, when it comes to RMC-C, in each iteration only two elements are selected for updating when fixing others, thereby resulting that some important manifolds like binary graph and cosine similarity graph are neglected by RMC-C during the optimization process. Different from RMC-C, RMC-E updates each element in a global view and has a global efficiency estimate for the convex minimization problem on the standard simplex.
\section{Conclusions}
\label{sec:conclusion}
This paper presents a novel co-clustering approach named \textit{Relational Multi-manifold Co-clustering} (RMC), which is based on the symmetric nonnegative matrix tri-factorization.

Our approach takes into account the inter-type relation and the intra-type information of both the sample and feature data simultaneously. The basic idea is to make use of manifold ensemble learning to enhance the performance of co-clustering. To achieve this, we attempt to learn a sensible convex combination of candidate manifolds so that it can maximally approximate the true intrinsic manifolds of both the sample and feature spaces. In order to optimize the objective function, we adopt the popular alternating optimization method to update the factorized matrices. However, different from the existing matrix factorization based co-clustering methods, there exists a manifold coefficient vector to be optimized in our approach, which poses a challenging task. In this work, we utilize two optimization methods with totally different mechanisms to optimize this coefficient vector, \textit{i.e.}, the \textit{entropic mirror descent algorithm} and the \textit{coordinate descent algorithm}. The effectiveness of the proposed approach is demonstrated by a number of interesting experiments on data collections from diverse domains, \textit{i.e.}, text processing, image analysis and bioinformatics.


%

%

\section*{Acknowledgment}
This work was supported in part by National Natural Science Foundation of China under Grants 91120302, 61222207, 60905001 and 61173186, National Basic Research Program of China (973 Program) under Grant 2013CB336500, the Fundamental Research Funds for the Central Universities under Grant 2012FZA5017 and the Zhejiang Province Key S$\&$T Innovation Group Project under Grant 2009R50009.

\bibliographystyle{IEEEtranS}
\bibliography{IEEEabrv,RMC}

\begin{thebibliography}{10}
\providecommand{\url}[1]{#1}
\csname url@samestyle\endcsname
\providecommand{\newblock}{\relax}
\providecommand{\bibinfo}[2]{#2}
\providecommand{\BIBentrySTDinterwordspacing}{\spaceskip=0pt\relax}
\providecommand{\BIBentryALTinterwordstretchfactor}{4}
\providecommand{\BIBentryALTinterwordspacing}{\spaceskip=\fontdimen2\font plus
\BIBentryALTinterwordstretchfactor\fontdimen3\font minus
  \fontdimen4\font\relax}
\providecommand{\BIBforeignlanguage}[2]{{%
\expandafter\ifx\csname l@#1\endcsname\relax
\typeout{** WARNING: IEEEtranS.bst: No hyphenation pattern has been}%
\typeout{** loaded for the language `#1'. Using the pattern for}%
\typeout{** the default language instead.}%
\else
\language=\csname l@#1\endcsname
\fi
#2}}
\providecommand{\BIBdecl}{\relax}
\BIBdecl

\bibitem{banerjee2004coclu-bregman}
A.~Banerjee, I.~Dhillon, J.~Ghosh, S.~Merugu, and D.~Modha, ``A generalized
  maximum entropy approach to bregman co-clustering and matrix approximation,''
  in \emph{Proceedings of the 10th ACM SIGKDD International Conference on
  Knowledge Discovery and Data Mining}, 2004, pp. 509--514.

\bibitem{beck2003mirror}
A.~Beck and M.~Teboulle, ``Mirror descent and nonlinear projected subgradient
  methods for convex optimization,'' \emph{Operations Research Letters},
  vol.~31, no.~3, pp. 167--175, 2003.

\bibitem{belkin2006manifold}
M.~Belkin, P.~Niyogi, and V.~Sindhwani, ``Manifold regularization: a geometric
  framework for learning from examples,'' \emph{Journal of Machine Learning
  Research}, vol.~7, pp. 2399--2434, 2006.

\bibitem{boyd2004convex}
S.~Boyd and L.~Vandenberghe, \emph{Convex optimization}.\hskip 1em plus 0.5em
  minus 0.4em\relax Cambridge University Press, 2004.

\bibitem{cai2005document-lpi}
D.~Cai, X.~He, and J.~Han, ``Document clustering using locality preserving
  indexing,'' \emph{{IEEE} Trans. Knowl. Data Eng.}, vol.~17, no.~12, pp.
  1624--1637, 2005.

\bibitem{cai2011gnmf}
D.~Cai, X.~He, J.~Han, and T.~Huang, ``Graph regularized nonnegative matrix
  factorization for data representation,'' \emph{{IEEE} Trans. Pattern Anal.
  Mach. Intell.}, vol.~33, no.~8, pp. 1548--1560, 2011.

\bibitem{cai2011lccf}
D.~Cai, X.~He, and J.~Han, ``Locally consistent concept factorization for
  document clustering,'' \emph{{IEEE} Trans. Knowl. Data Eng.}, vol.~23, no.~6,
  pp. 902--913, 2011.

\bibitem{cai2009lpnmf}
D.~Cai, X.~He, X.~Wang, H.~Bao, and J.~Han, ``Locality preserving nonnegative
  matrix factorization,'' in \emph{Proceedings of the International Joint
  Conference on Artificial Intelligence}, 2009, pp. 1010--1015.

\bibitem{cai2008nmf-manifold}
D.~Cai, X.~He, X.~Wu, and J.~Han, ``Non-negative matrix factorization on
  manifold,'' in \emph{Proceedings of the 8th IEEE International Conference on
  Data Mining}, 2008, pp. 63--72.

\bibitem{cai2009dyadic}
D.~Cai, X.~Wang, and X.~He, ``Probabilistic dyadic data analysis with local and
  global consistency,'' in \emph{Proceedings of the 26th International
  Conference On Machine Learning}, 2009, pp. 105--112.

\bibitem{chen2009cf-onmtf}
G.~Chen, F.~Wang, and C.~Zhang, ``Collaborative filtering using orthogonal
  nonnegative matrix tri-factorization,'' \emph{Information Processing $\&$
  Management}, vol.~45, no.~3, pp. 368--379, 2009.

\bibitem{chen2007simulcluster}
S.~Chen, F.~Wang, and C.~Zhang, ``Simultaneous heterogeneous data clustering
  based on higher order relationships,'' in \emph{Proceedings of the 7th IEEE
  International Conference on Data Mining}, 2007, pp. 387--392.

\bibitem{chen2011parallel-sc}
W.~Chen, Y.~Song, H.~Bai, C.~Lin, and E.~Chang, ``Parallel spectral clustering
  in distributed systems,'' \emph{{IEEE} Trans. Pattern Anal. Mach. Intell.},
  vol.~33, no.~3, pp. 568--586, 2011.

\bibitem{chen2010nmf-semi-coclu}
Y.~Chen, L.~Wang, and M.~Dong, ``Non-negative matrix factorization for
  semisupervised heterogeneous data coclustering,'' \emph{{IEEE} Trans. Knowl.
  Data Eng.}, vol.~22, no.~10, pp. 1459--1474, 2010.

\bibitem{cho2004coclu-gene}
H.~Cho, I.~Dhillon, Y.~Guan, and S.~Sra, ``Minimum sum-squared residue
  co-clustering of gene expression data,'' in \emph{Proceedings of the 4th SIAM
  International Conference on Data Mining}, vol. 114, 2004, pp. 114--125.

\bibitem{dai2007coclu-doc}
W.~Dai, G.~Xue, Q.~Yang, and Y.~Yu, ``Co-clustering based classification for
  out-of-domain documents,'' in \emph{Proceedings of the 13th ACM SIGKDD
  International Conference on Knowledge Discovery and Data Mining}, 2007, pp.
  210--219.

\bibitem{dhillon2001coclu-bipartite}
I.~Dhillon, ``Co-clustering documents and words using bipartite spectral graph
  partitioning,'' in \emph{Proceedings of the 9th ACM SIGKDD International
  Conference on Knowledge Discovery and Data Mining}, 2001, pp. 269--274.

\bibitem{dhillon2003coclu-info}
I.~Dhillon, S.~Mallela, and D.~Modha, ``Information-theoretic co-clustering,''
  in \emph{Proceedings of the 9th ACM SIGKDD International Conference on
  Knowledge Discovery and Data Mining}, 2003, pp. 89--98.

\bibitem{ding2006onmtf}
C.~Ding, T.~Li, W.~Peng, and H.~Park, ``Orthogonal nonnegative matrix
  t-factorizations for clustering,'' in \emph{Proceedings of the 12th ACM
  SIGKDD International Conference on Knowledge Discovery and Data Mining},
  2006, pp. 126--135.

\bibitem{ding2010convex}
C.~Ding, T.~Li, and M.~Jordan, ``Convex and semi-nonnegative matrix
  factorizations,'' \emph{{IEEE} Trans. Pattern Anal. Mach. Intell.}, vol.~32,
  no.~1, pp. 45--55, 2010.

\bibitem{geng2009emr}
B.~Geng, C.~Xu, D.~Tao, L.~Yang, and X.~Hua, ``Ensemble manifold
  regularization,'' in \emph{Proceedings of the IEEE Conference on Computer
  Vision and Pattern Recognition}, 2009, pp. 2396--2402.

\bibitem{george2005cf-coclu}
T.~George and S.~Merugu, ``A scalable collaborative filtering framework based
  on co-clustering,'' in \emph{Proceedings of the IEEE International Conference
  on Data Mining}, 2005, pp. 625--628.

\bibitem{gu2009coclu-graph}
Q.~Gu and J.~Zhou, ``Co-clustering on manifolds,'' in \emph{Proceedings of the
  15th ACM SIGKDD International Conference on Knowledge Discovery and Data
  Mining}, 2009, pp. 359--367.

\bibitem{hanisch2002coclu-bio}
D.~Hanisch, A.~Zien, R.~Zimmer, and T.~Lengauer, ``Co-clustering of biological
  networks and gene expression data,'' \emph{Bioinformatics}, vol.~18, no.~S1,
  pp. S145--S154, 2002.

\bibitem{khan2001srbct}
J.~Khan, J.~Wei, M.~Ringn{\'e}r, L.~Saal, M.~Ladanyi, F.~Westermann,
  F.~Berthold, M.~Schwab, C.~Antonescu, C.~Peterson \emph{et~al.},
  ``Classification and diagnostic prediction of cancers using gene expression
  profiling and artificial neural networks,'' \emph{Nature medicine}, vol.~7,
  no.~6, pp. 673--679, 2001.

\bibitem{lee1999nmf}
D.~Lee, H.~Seung \emph{et~al.}, ``Learning the parts of objects by non-negative
  matrix factorization,'' \emph{Nature}, vol. 401, no. 6755, pp. 788--791,
  1999.

\bibitem{lewis2004rcv1}
D.~Lewis, Y.~Yang, T.~Rose, and F.~Li, ``Rcv1: A new benchmark collection for
  text categorization research,'' \emph{Journal of Machine Learning Research},
  vol.~5, pp. 361--397, 2004.

\bibitem{liu2008alphagene}
W.~Liu, K.~Yuan, and D.~Ye, ``On $\alpha$-divergence based nonnegative matrix
  factorization for clustering cancer gene expression data,'' \emph{Artificial
  Intelligence in Medicine}, vol.~44, no.~1, pp. 1--5, 2008.

\bibitem{long2006sc-multirelation}
B.~Long, Z.~Zhang, X.~Wu, and P.~Yu, ``Spectral clustering for multi-type
  relational data,'' in \emph{Proceedings of the 23rd International Conference
  on Machine Learning}, 2006, pp. 585--592.

\bibitem{long2005coclu-bvd}
B.~Long, Z.~Zhang, and P.~Yu, ``Co-clustering by block value decomposition,''
  in \emph{Proceedings of the 11th ACM SIGKDD International Conference on
  Knowledge Discovery and Data Mining}, 2005, pp. 635--640.

\bibitem{pan2008crd}
F.~Pan, X.~Zhang, and W.~Wang, ``Crd: Fast co-clustering on large datasets
  utilizing sampling-based matrix decomposition,'' in \emph{Proceedings of the
  ACM SIGMOD International Conference on Management of Data}, 2008, pp.
  173--184.

\bibitem{platt1999smo}
J.~Platt, ``Fast training of support vector machines using sequential minimal
  optimization,'' in \emph{Advances in Kernel Methods}.\hskip 1em plus 0.5em
  minus 0.4em\relax MIT Press, 1999.

\bibitem{rege2006coclu-bipartie}
M.~Rege, M.~Dong, and F.~Fotouhi, ``Co-clustering documents and words using
  bipartite isoperimetric graph partitioning,'' in \emph{Proceedings of the
  IEEE International Conference on Data Mining}, 2006, pp. 532--541.

\bibitem{sindhwani2008regu-coclu}
V.~Sindhwani, J.~Hu, and A.~Mojsilovic, ``Regularized co-clustering with dual
  supervision,'' in \emph{Advances in Neural Information Processing Systems},
  vol.~21, 2008, pp. 1505--1512.

\bibitem{slonim2000mutualinfo}
N.~Slonim and N.~Tishby, ``Document clustering using word clusters via the
  information bottleneck method,'' in \emph{Proceedings of the 23rd ACM SIGIR
  Conference on Research and Development in Information Retrieval}.\hskip 1em
  plus 0.5em minus 0.4em\relax ACM, 2000, pp. 208--215.

\bibitem{statnikov2005gene-bio}
A.~Statnikov, C.~Aliferis, I.~Tsamardinos, D.~Hardin, and S.~Levy, ``A
  comprehensive evaluation of multicategory classification methods for
  microarray gene expression cancer diagnosis,'' \emph{Bioinformatics},
  vol.~21, no.~5, pp. 631--643, 2005.

\bibitem{wang2008ssc-nmf}
F.~Wang, T.~Li, and C.~Zhang, ``Semi-supervised clustering via matrix
  factorization,'' in \emph{Proceedings of the 8th SIAM International
  Conference on Data Mining}, vol.~1, 2008, pp. 1--12.

\bibitem{wang2011snmtf}
H.~Wang, H.~Huang, and C.~Ding, ``Simultaneous clustering of multi-type
  relational data via symmetric nonnegative matrix tri-factorization,'' in
  \emph{Proceedings of the 20th ACM International Conference on Information and
  Knowledge Management}.\hskip 1em plus 0.5em minus 0.4em\relax ACM, 2011, pp.
  279--284.

\bibitem{wang2011osntf}
H.~Wang, F.~Nie, H.~Huang, and C.~Ding, ``Nonnegative matrix tri-factorization
  based high-order co-clustering and its fast implementation,'' in
  \emph{Proceedings of the IEEE International Conference on Data Mining}, 2011,
  pp. 774--783.

\bibitem{wang2011fast-nmtf}
H.~Wang, F.~Nie, H.~Huang, and F.~Makedon, ``Fast nonnegative matrix
  tri-factorization for large-scale data co-clustering,'' in \emph{Proceedings
  of the 22nd IEEE International Joint Conference on Artificial Intelligence},
  vol.~2, 2011, pp. 1553--1558.

\bibitem{xu2006pattern-coclu}
X.~Xu, Y.~Lu, A.~Tung, and W.~Wang, ``Mining shifting-and-scaling co-regulation
  patterns on gene expression profiles,'' in \emph{Proceedings of the 22nd IEEE
  International Conference on Data Engineering}.\hskip 1em plus 0.5em minus
  0.4em\relax IEEE, 2006, pp. 89--89.

\bibitem{zha2001bipartite}
H.~Zha, X.~He, C.~Ding, H.~Simon, and M.~Gu, ``Bipartite graph partitioning and
  data clustering,'' in \emph{Proceedings of the 10th ACM International
  Conference on Information and Knowledge Management}.\hskip 1em plus 0.5em
  minus 0.4em\relax ACM, 2001, pp. 25--32.

\bibitem{zhang2012ldcc}
L.~Zhang, C.~Chen, J.~Bu, Z.~Chen, D.~Cai, and J.~Han, ``Locally discriminative
  co-clustering,'' \emph{{IEEE} Trans. Knowl. Data Eng.}, vol.~24, no.~6, pp.
  1025--1035, 2012.

\bibitem{zhi2011gsnmf}
R.~Zhi, M.~Flierl, Q.~Ruan, and W.~Kleijn, ``Graph-preserving sparse
  nonnegative matrix factorization with application to facial expression
  recognition,'' \emph{{IEEE} Trans. Syst., Man, Cybern. {B}}, vol.~41, no.~1,
  pp. 38--52, 2011.

\end{thebibliography}

\ifCLASSOPTIONcaptionsoff
  \newpage
\fi

\end{document}